\documentclass[journal]{IEEEtran}
\ifCLASSOPTIONcompsoc
  \usepackage[nocompress]{cite}
\else
  \usepackage{cite}
\fi

\usepackage{amsmath,amsfonts}
\usepackage{algorithmic}
\usepackage[linesnumbered,ruled,vlined]{algorithm2e}
\usepackage{array}
\usepackage[caption=false,font=footnotesize,labelfont=sf,textfont=sf]{subfig}
\usepackage{textcomp}
\usepackage{stfloats}
\usepackage{url}
\usepackage{verbatim}
\usepackage{graphicx}
\usepackage{cite}
\usepackage{utfsym}
\usepackage{multirow}
\usepackage{booktabs}
\usepackage[misc]{ifsym}
\usepackage{ntheorem}
\usepackage{float}
\usepackage{threeparttable}
\usepackage{caption}
\usepackage{balance}
\usepackage{colortbl}
\usepackage{diagbox}
\definecolor{mygray}{gray}{.7}
\newtheorem{definition}{Definition}
\newtheorem{lemma}{Lemma}
\newtheorem*{proof}{Proof}
\usepackage{setspace}
\usepackage{float}
\usepackage{titlesec}

\def\ie{{\em i.e.}}
\def\eg{{\em e.g.}}
\def\etal{{\em et al.}}
\newtheorem{theorem}{Theorem}{}

{}

\newcommand{\bl}[1]{\textbf{#1}}

\newcommand{\mc}[1]{\mathcal{#1}}
\newcommand{\mb}[1]{\mathbb{#1}}

\graphicspath{{./fig/}}

\newcommand{\myPara}[1]{\vspace{.05in}\noindent\textbf{#1}}

\begin{document}
\title{Privacy-Preserving Model Transcription with Differentially Private Synthetic Distillation}
\author{Bochao Liu, Shiming Ge, \IEEEmembership{Senior Member, IEEE}, Pengju Wang, Shikun Li, and Tongliang Liu, \IEEEmembership{Senior Member, IEEE}%
\IEEEcompsocitemizethanks{\IEEEcompsocthanksitem Bochao Liu, Shiming Ge, Pengju Wang and Shikun Li are with the Institute of Information Engineering at Chinese Academy of Sciences, Beijing 100085, China. E-mail: \{liubochao, geshiming, wangpengju, lishikun\}@iie.ac.cn. \IEEEcompsocthanksitem Bochao Liu is also with the Beijing Institute of Astronautical Systems Engineering, 100076 Beijing, China. \IEEEcompsocthanksitem Tongliang Liu is with Trustworthy Machine Learning Lab, School of Computer Science, The University of Sydney, Camperdown, NSW 2050, Australia. E-mail: tongliang.liu@sydney.edu.au.}
\thanks{Shiming Ge is the corresponding author. E-mail: geshiming@iie.ac.cn.}
}

\markboth{Submitted to IEEE Transactions on Pattern Analysis and Machine Intelligence}%
{Shell \MakeLowercase{\textit{et al.}}: Bare Demo of IEEEtran.cls for Computer Society Journals}

\maketitle
\begin{abstract}
While many deep learning models trained on private datasets have been deployed in various practical tasks, they may pose a privacy leakage risk as attackers could recover informative data or label knowledge from models. In this work, we present \emph{privacy-preserving model transcription}, a data-free model-to-model conversion solution to facilitate model deployment with a privacy guarantee. To this end, we propose a cooperative-competitive learning approach termed \emph{differentially private synthetic distillation} that learns to convert a pretrained model (teacher) into its privacy-preserving counterpart (student) via a trainable generator without access to private data. The learning collaborates with three players in a unified framework and performs alternate optimization: i)~the generator is learned to generate synthetic data, ii)~the teacher and student accept the synthetic data and compute differential private labels by flexible data or label noisy perturbation, and iii)~the student is updated with noisy labels and the generator is updated by taking the student as a discriminator for adversarial training. We theoretically prove that our approach can guarantee differential privacy and convergence. The transcribed student has good performance and privacy protection, while the resulting generator can generate private synthetic data for downstream tasks. Extensive experiments clearly demonstrate that our approach outperforms 26 state-of-the-arts. 
\end{abstract}


\maketitle

\section{Introduction}
\begin{figure}[!t]
\begin{center}
\includegraphics[width=1.0\linewidth]{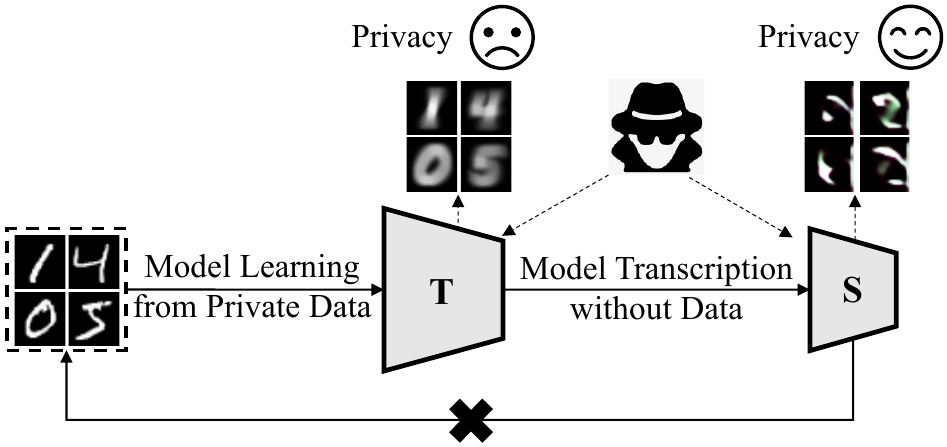}
\caption{A model \textbf{T} directly trained on private data may leak privacy in practical deployment, which inspires us to propose {model transcription} that can convert \textbf{T} into a privacy-preserving model \textbf{S} without access to private data.}
\label{fig:motivation}
\vspace{-10pt}
\end{center}
\end{figure}

\IEEEPARstart{D}{eep} learning models~\cite{krizhevsky2012imagenet,vggnet2015iclr,resnet2016cvpr,dosovitskiy2020image} are widely used to learn rich knowledge from substantial real-world datasets, subsequently deployed in various critical tasks. 
For instance, a three-layer fully-connected net was trained on the extensive binary classification dataset SocialEvent for abnormal detection via trustable co-label learning and gave a validate AUC of 75.21\%~\cite{li2023tmm}, while a ResNet64 model was trained on massive personal identity photos for ID versus Spot face recognition via large-scale bisample learning and achieved a verification rate of 96.43\% at low false acceptance rate~\cite{zhu2019large}. 
Beyond their success, there is growing concern about their potential to compromise privacy from training data. This could occur through model inversion attacks with limited access~\cite{fredrikson2015ccs,yang2019ccs} or privacy violation via inference with large language models~\cite{openai2023gpt4,touvron2023llama}, especially as these models become accessible~\cite{staab2023memorization}, as shown in Fig.~\ref{fig:motivation}. Therefore, high-accuracy models with a privacy guarantee are critical for real-world deployment. To this end, exploring a feasible solution that can address a key challenge in model re-deployment is necessary: {how to effectively convert a well-trained model into a privacy-preserving one without significant loss of inference accuracy?}

An intuitive idea for model re-deployment is model compression \cite{deng2020pieee} whose main objective is to reduce model parameters or storage so that the model can be deployed into resource-limited devices. Existing approaches primarily rely on network quantization, network pruning, and knowledge distillation. Network quantization and pruning reduce the model parameters by low-precision approximation~\cite{tung2020tpami,wang2021tnnls} and redundancy removal~\cite{liu2022tpami,yvinec2023tpami}, respectively, which mitigates model sensitivity to small data change. In contrast, knowledge distillation~\cite{gou2021tpami,rishav2021kcd,faisal2023privacy} involves using a pretrained model as teacher to train a more compact student model by behavior mimicking. Such solution allows the distilled student to avoid direct access to private data by training on public~\cite{wang2019aaai} or generated data~\cite{lopes2017data,chen2019data}. However, evaluating the privacy protection capability of these model compression techniques is challenging. A common strategy for privacy evaluation is differential privacy~\cite{dwork2016jpc}, which aims to learn models on private data while protecting data privacy~\cite{abadi2016deep} and has been modified to achieve label privacy~\cite{chaudhuri2011sample,ghazi2021nips,esfandiari2022aistats}. Nonetheless, when private data are unavailable, applying differential privacy directly to learn private models is impractical. To overcome this, some works~\cite{papernot2016semi,jordon2018pate,ge2023tip} have integrated knowledge distillation with differential privacy, using an ensemble of pretrained models as teachers to learn private models on public data. A recent model conversion approach~\cite{liu2023model} generated synthetic data and then exploited them to distill pretrained models into privacy-preserving ones without access to private data. These approaches demonstrate the potential of combining knowledge distillation and differential privacy for private model training without original datasets but struggle to distill the rich knowledge from teachers to students with a privacy guarantee. The challenge lies in finding a reasonable and effective trade-off between model performance and privacy protection, which involves addressing three main issues: i) \emph{high privacy cost} for high-dimensional data, ii) \emph{blind synthetic data generation} with representation distribution of private data, and iii) \emph{priority selection} to meet practical data-sensitive or label-sensitive privacy requirements.
   
\begin{figure*}[t]
\centering
\includegraphics[width=1.0\linewidth]{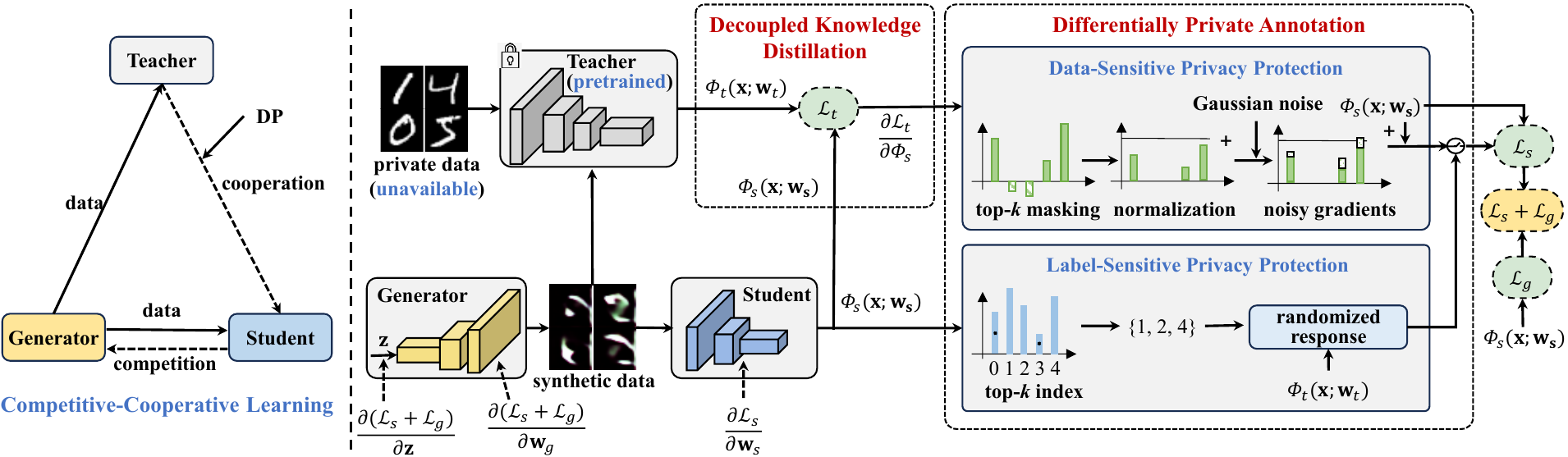}
\caption{\textbf{Overview of our differentially private synthetic distillation approach.} Left: Our approach can transcribe a pretrained \textbf{teacher} into a privacy-preserving \textbf{student} without available private data via a trainable \textbf{generator} by competitive-cooperative learning. It collaborates with three players in a unified framework, where the student takes part in cooperation with the teacher under differential privacy and competition with the generator. Right: The competitive-cooperative learning freezes the teacher, initializes the student and generator, and then alternately updates the student and generator in each iteration round. First, the generator generates synthetic data from random vectors $\bl{z}$ under Gaussian distribution. After that, the student predictions $\Phi_s(\bl{x};\bl{w}_s)$ and teacher predictions $\Phi_t(\bl{x};\bl{w}_t)$ are used to perform differentially private annotation to the synthetic data, which provides switchable data-sensitive or label-sensitive privacy protection with normalization as well as Gaussian mechanism or randomized response mechanism, respectively. Specially, top-$k$ scheme for value masking or index selection is used to reduce the privacy budget and improve model accuracy. Finally, the student and generator are alternately updated by decoupled knowledge distillation with noisy annotations and adversarial learning with the student as a discriminator, respectively.}
\label{fig:framework}
\end{figure*}

To address these challenges, this paper proposes a differentially private synthetic distillation approach to facilitate privacy-preserving model transcription (Fig.~\ref{fig:motivation}). This approach aims to convert a pretrained model (teacher) into a privacy-preserving one (student) without available private data. It takes a trainable generator as bridge and engages three players in a competitive-cooperative learning framework (Fig.~\ref{fig:framework}), where the student collaborates with the teacher through knowledge distillation and competes with the generator via adversarial learning. Through alternate optimization of synthetic data generation and model training, the competitive-cooperative learning process achieves convergence when the student mimics the teacher in inference capability, and the generator produces synthetic data that match the representation distribution of private data. In the learning framework, the teacher and student collaborate to efficiently annotate synthetic data in a differentially private manner, facilitating the transfer of teacher knowledge to the student and effectively mitigating the issue of \emph{high privacy cost}. Despite the absence of private data, the student inherits the teacher's ability to discriminate on private data and acts as a discriminator in competition with the generator. This setup makes \emph{blind synthetic data generation} effective, as the student's refined discriminative capability guides the generation of synthetic data that closely resemble private data distribution. Further, our approach offers switchable \emph{priority selection} in a unified framework to protect either data-sensitive privacy via Gaussian mechanism or label-sensitive privacy via randomized response mechanism, catering to specific privacy needs (\eg, in preference to protecting facial images or identities). {To this end, we design theoretical differentially private annotation to balance model performance and privacy protection by i) gradient normalization in Gaussian mechanism to retain relative scale information, ii) utilizing student outputs in randomized response mechanism to retain more information and maintain privacy integrity, and iii) exploiting lower-dimensional labels to achieve differential privacy.} In this way, we can theoretically validate both privacy and convergence guarantee of our approach, achieving a privacy-preserving student with minimal accuracy loss. 

Our major contributions are three folds: 1) we propose a differentially private synthetic distillation approach for model transcription without available private data, 2) we propose a unified framework to support two flexible priority selection in protecting data or label privacy under guarantees in theory, and 3) we conduct extensive experiments and analysis to demonstrate the effectiveness of our approach.
    
\section{Related Works}
\myPara{Model privacy protection.}~Many works have been proposed to protect the privacy of released models through various techniques. Model watermarking framework~\cite{zhang2022tpami} was proposed to actively protect the intellectual property of networks, while some approaches applied machine unlearning techniques to passively address model privacy~\cite{bourtoule2021sp,xu2024csur}. Another idea is model conversion via knowledge distillation~\cite{gou2021tpami,rishav2021kcd,faisal2023privacy}. To further avoid direct data access, some approaches employed data-free knowledge distillation~\cite{chen2019data,fang2022up} to safeguard the models. Some researchers protect models by training them with desensitized data~\cite{wang2021ccs,dockhorn2023tmlr,chen2022dpgen}. For example, Chen \etal~\cite{chen2020gs} used Wasserstein GANs~\cite{arjovsky2017wasserstein} for a precise estimation of the sensitivity value, avoiding the intensive search of hyper-parameters while reducing the clipping bias. Recent diffusion model~\cite{ho2020denoising} was used in \cite{dockhorn2023tmlr,liu2024tmlr,ghalebikesabi2023differentially} for training private models on large datasets. Without differential privacy (DP)~\cite{dwork2016jpc}, evaluating privacy protection ability of these approaches is difficult, but there is still a long way to get a good balance between model utility and privacy.

\myPara{Differentially private learning.}~Its objective is to ensure model learning is differentially private regarding the private data. To achieve DP, the early work \cite{abadi2016deep} proposed differentially private stochastic gradient descent~(DPSGD) by adding noise to the gradients of all training parameters, and later this idea was extended by many works~\cite{xie2018differentially,chen2020gs,cao2021don}. Typically, the DPSGD-based approaches can meet strong privacy requirements but seriously degrade model performance. To improve private learning, Papernot \etal~\cite{papernot2016semi} proposed private aggregation of teacher ensembles (PATE) by transferring knowledge of public data, and then some works~\cite{papernot2018scalable,bassily2018model,duan2023flocks} followed PATE idea where the key challenge is to achieve training data (\eg, from public) that have similar representation distribution to private data. To this end, recent approaches~\cite{wang2021ccs,long2021g,jordon2018pate,chen2022dpgen,liu2023model} learned privacy-preserving generative models to generate synthetic training data. For example, Chen \etal~\cite{chen2022dpgen} proposed an energy-guided network trained on sanitized data to improve data generation, while Liu \etal~\cite{liu2023model} combined differential privacy and data generation to achieve private model transcription. In general, the results of these approaches are not satisfactory, and the key to balance privacy protection and model performance needs to reliable data generation and effective knowledge transfer, which inspires our approach.    

\section{The Proposed Approach}
Without available private data, we take an alternative solution to perform model-to-model conversion by introducing a trainable generator into the teacher-student learning framework. Thus, the generator should improve student learning by meeting \emph{two rules}: representation distribution similarity between generated and private data, and privacy protection of private data. Inspired by that, our differentially private synthetic distillation (DPSD) approach involves three players (see Fig. \ref{fig:framework}), including the teacher, the student and the generator. In this way, the rich private knowledge contained in the teacher can be distilled and transferred to guide the process of data generation and student learning. 

\subsection{Overview}
\emph{1) Problem Statement.} Our objective is transcribing a pretrained model $\Phi_t(\bl{x};\bl{w}_t)$ (teacher) into its privacy-preserving counterpart $\Phi_s(\bl{x};\bl{w}_s)$ (student), where $\bl{x}$ is an input example during deployment, $\bl{w}_t$ and $\bl{w}_s$ are the model parameters, respectively. Generally, the private dataset $\hat{\mc{D}}$ used to train $\Phi_t$ is not available, and using public datasets for training models often leads to an accuracy decrease due to representation distribution discrepancy~\cite{chen2019data,ge2023tip}. Thus,  a trainable generator $\Phi_g$ is introduced to generate synthetic image $\bl{x}=\Phi_g(\bl{z};\bl{w}_g)$ from a noisy vector $\bl{z}$, where $\bl{w}_g$ are the generator parameters. Then, the private learning is performed by the following:
\begin{subequations}
\begin{align}
\Pi[\Phi_g(\bl{z};\bl{w}_g)]=\Pi[\hat{\mc{D}}],\\
\Phi_s(\bl{x};\bl{w}_s)\doteq\Phi_t(\bl{x};\bl{w}_t),
\end{align}
\end{subequations}
where $\Pi[*]$ measures the representation distribution of data, and $\doteq$ means ``equivalence'' in some metric, \eg, classification probability. Eq.(1a) tries to generate data to meet the first rule by reducing representation distribution discrepancy between generated and private data, while Eq.(1b) enforces the same capacity between the student and teacher under the privacy guarantee to meet the second rule. Now, the question is how to measure the two rules and how to enforce the ``equivalence'' with privacy guarantee. Inspired by that, we propose differentially private synthetic distillation (DPSD).

\emph{2) DPSD.} It involves three players in a unified cooperation-competition learning framework. The teacher $\Phi_t$ pretrained on private data (generally unavailable) is frozen and transfers knowledge in cooperation with the student. The student $\Phi_s$ is learned to mimic the teacher's behaviors and serves as a discriminator in competition with the generator. The generator $\Phi_g$ is learned to generate synthetic data that have the same representation distribution as private data. By alternate synthetic data generation and learning on them, the cooperation-competition learning converges when the student behaves the same as the teacher under differential privacy, as well as the generator, produces synthetic data that have the same representation distribution as private data. Our approach takes the student rather than the teacher in other approaches~\cite{chen2019data,ge2023tip} as discriminator for generator learning, making the learned generator also achieve differential privacy.

\subsection{Differentially Private Synthetic Distillation}
Given a fixed teacher $\Phi_t$, DPSD first initializes the generator $\Phi_g$ and the student $\Phi_s$ via Xavier method, randomly initializes Gaussian noise vectors, and then learns them in an alternate optimization manner. In each iteration round, the alternate learning consists of four main steps, including synthetic data generation, differentially private annotation, student update via distillation, and generator update via adversarial learning.

\emph{1) Synthetic Data Generation.} This step samples $b$ Gaussian noise vectors $\{\bl{z}_i\}_{i=1}^b$ and feeds them into $\Phi_g$ to require a batch of synthetic data $\{\bl{x}_i\}_{i=1}^{b}$, where $\bl{x}_i=\Phi_g(\bl{z}_i;\bl{w}_g)$. Subsequently, the synthetic data can be fed into $\Phi_t$ and $\Phi_s$ to obtain their predictions $\Phi_t(\bl{x}_i;\bl{w}_t)$ and $\Phi_s(\bl{x}_i;\bl{w}_s)$. To enforce data generation, Gaussian noise vectors are involved for updating in the learning. 

\emph{2) Differentially Private Annotation.} To achieve differential privacy, our approach provides switchable annotation for selectively protecting data- or label-sensitive privacy. The annotation is implemented using differential privacy tailored to the teacher and student predictions. For an example $\bl{x}_i$, we treat the teacher prediction as its target label $r=\arg\max_j\Phi_t(\bl{x}_i;\bl{w}_t)_j$, and the student prediction as prior knowledge to form an index set $\mc{I}_i=\lceil\Phi_s(\bl{x}_i;\bl{w}_s)\rceil_k$ by comprising the top-$k$ most probable classes, where the operator $\lceil*\rceil_k$ determines the index set corresponding to top-$k$ largest values of a vector, where $2 \le k\le c$ and $c$ is class number. We denote an operator $\mc{M}_k(*)$ to keep the top-$k$ largest elements of a vector and set others to zeros.

In data-sensitive privacy protection, we denote $p_{*,i,j}=\Phi_*(\bl{x}_i;\bl{w}_t)_{j}, *=\{t,s\}$ for simplicity, and subsequently calculate decoupled knowledge distillation loss~\cite{zhao2022decoupled} that consists of target class loss $\ell_{tc}$ and non-target class loss $\ell_{nc}$:  
\begin{equation}\label{eq:LT}
\begin{aligned}
\mc{L}_t(\Phi_t,\Phi_s, \{\bl{x}_i\}_{i=1}^b)=\sum_{i=1}^b (\ell_{tc}(\bl{x}_i) + \lambda\ell_{nc}(\bl{x}_i)),
\end{aligned}
\end{equation}
where $\lambda$ is used to balance the effect of two losses and we set it as 8.0 in our experiments, 
$\ell_{tc}(\bl{x}_i)=p_{t,i,r}\log\frac{p_{t,i,r}}{p_{s,i,r}}+(1-p_{t,i,r})\log\frac{1-p_{t,i}}{1-p_{s,i,r}}$ and
$\ell_{nc}(\bl{x}_i)=\sum_{j\neq r}\frac{p_{t,i,j}}{1-p_{t,i,j}}\log\frac{p_{t,i,j}(1-p_{s,i,j})}{p_{s,i,j}(1-p_{t,i,j})}$.
Then, we compute the gradients, normalize them to limit the sensitivity, subsequently incorporate the Gaussian mechanism to generate differentially private gradients, and finally achieve the noisy annotation:
\begin{equation}\label{eq:dp}
\begin{aligned}
\bl{y}_i^{(d)} &= \Phi_s(\bl{x}_i;\bl{w}_s) \\
&-\frac{\gamma_s}{b}\sum\limits^{b}\limits_{i=1}(\frac{\beta\cdot {(\mc{M}_k(\partial\mc{L}_t}/{\partial\Phi_s}))_i}{||\mc{M}_k({\partial\mc{L}_t}/{\partial\Phi_s})||_2+h} + \mc{N}(0,\sigma^2\beta^2)),
\end{aligned}
\end{equation}
where $\gamma_s$ is the learning rate, $h$ is a stability constant and set to $10^{-4}$ for avoiding zero denominators, $\mc{N}$ is Gaussian distribution whose mean is 0 and noise variance is controlled by the noise scale $\sigma$ and normalization bound $\beta$. The noise scale affects the privacy budget $\varepsilon$.

In label-sensitive privacy protection, we use randomized response mechanism $\mb{R}$~\cite{warner1965randomized} to achieve label differential privacy (LabelDP)~\cite{ghazi2021nips}  by defining the probabilities of returning correct class label $r$ and other labels: 
\begin{equation}\label{eq:rr}
\operatorname{Pr}[\mb{R}(r; c, \varepsilon)={y}]= \begin{cases}{e^{\varepsilon}}/{(e^{\varepsilon}+c-1)}, & \text { if } y=r \\ 1/{(e^{\varepsilon}+c-1)}, & \text { otherwise } \end{cases},
\end{equation}
where $\varepsilon$ is privacy budget, $y\in\{1,2,...,c\}$. To enhance the likelihood of returning correct class labels without compromising privacy, we incorporate the student predictions as additional information. Specifically, a top-$k$ selection is conducted using the student outputs as prior knowledge and an index set $\mc{I}_i$ is formed for $\bl{x}_i$. Thus, the annotation for label-sensitive privacy can be formulated as: 
\begin{equation}\label{eq:labelprivacy}
\bl{y}_i^{(l)}= \begin{cases}\mc{H}[\mb{R}(r; c, \varepsilon)], & \text { if } r\in{\mc{I}_i}\\ \mc{H}[\mb{U}({\lceil\Phi_s(\bl{x}_i;\bl{w}_s)\rceil_k})], & \text { otherwise }\end{cases},
\end{equation}
where the operator $\mc{H}$ is used to convert a class label into a one-hot vector, $\mb{U}$ is a uniform sampling function to return any element in $\mc{I}_i$ randomly with probability $1/k$. Overall, the label-sensitive privacy protection can be formally proved LabelDP in Theorem~\ref{th:theorem-label}.

From Eq. \eqref{eq:dp} and Eq. \eqref{eq:labelprivacy}, the annotation process will perturb the knowledge for preserving privacy according to the privacy selection, which can be rewritten in a unified way by introducing a binary switch sign $s\in\{0,1\}$: 
\begin{equation}\label{eq:noisyannotation}
\begin{aligned}
\hat{\bl{y}}_i = s\bl{y}_i^{(d)}+(1-s)\bl{y}_i^{(l)}.
\end{aligned}
\end{equation}

\emph{3) Student Update via Distillation.} After differentially private annotation, the noisy labels can be seen as soft labels. Using the batch of data $\{\bl{x}_i,\hat{\bl{y}}_i\}_{i=1}^b$ with soft targets, the student is updated by minimizing the loss $\mc{L}_s$:
\begin{equation}\label{eq:ls}
\begin{aligned}
\mc{L}_s(\{\bl{x}_i,\hat{\bl{y}}_i\}_{i=1}^b;\bl{w}_s)=\sum_{i=1}^b\ell(\Phi_s(\bl{x}_i;\bl{w}_s),\hat{\bl{y}}_i),
\end{aligned}
\end{equation}
where $\ell(\cdot)$ measures the cross-entropy loss. In this way, the student can mimic the behaviors of the teacher, leading to effective knowledge transfer~\cite{hinton2015distilling,zhao2022decoupled}. 
    
\emph{4) Generator Update via Adversarial Learning.} By learning knowledge from the teacher, the student is updated to have a similar discriminative ability to private data. Thus, we can treat the student as a discriminator variant and enable generator update via adversarial learning. Here, employing a multi-class classifier as the discriminator, instead of a binary classifier, has been shown to be more effective in learning representation distributions~\cite{chen2019data, ge2023tip}. Inspired by that, we incorporate an additional loss function $\mc{L}_g$ alongside $\mc{L}_s$ when updating the generator $\Phi_g$, formulated as,
\begin{equation}\label{eq:generatorupdate}
\begin{aligned}
\mc{L}_g(\{\bl{x}_i\}_{i=1}^b)=&\sum_{i=1}^b\ell(\Phi_s(\bl{x}_i;\bl{w}_s),\arg\max_j\Phi_s(\bl{x}_i)_j)\\
+&\sum_{i=1}^b\Phi_s(\bl{x}_i;\bl{w}_s)\log(\Phi_s(\bl{x}_i;\bl{w}_s))\\+&\sum_{i=1}^b||\Phi_s(\bl{x}_i;\bl{w}_s^{-})||_2,
\end{aligned}
\end{equation}  
where $\bl{x}_i=\Phi_g(\bl{z}_i;\bl{w}_g)$ contains the trainable generator parameters $\bl{w}_g$, $\bl{w}_s^{-} \subset \bl{w}_s$ are the parameters of student backbone. The first term measures one-hot classification loss, which can enforce synthetic data having the same representation distribution as private data. The second term is information entropy loss, which measures the class balance of generated data. The third term measures activation loss with $l_2$-norm, since the features that are extracted by the student and correspond to the output before the fully-connected layers tend to receive some higher activation values if input data are real rather than some random vectors. 

Alg.~\ref{alg:DPSD} shows the algorithm detail, where differentially private annotation refers to selective protection of data- or label-sensitive privacy. With differentially private annotations, the trained student and generator essentially satisfy differential privacy (DP)~\cite{dwork2016jpc} or label differential privacy (LabelDP)~\cite{ghazi2021nips} since the training can be seen as post-processing~\cite{dwork2014algorithmic}. To speed up, we include the generator inputs into the training. In each round, we update the generator as well as its inputs with $\mc{L}_s+\mc{L}_g$ shown in Fig.~\ref{fig:framework} and sample data with the trained inputs to update the student with $\mc{L}_s$. Compared to directly training the generator and then fixing it to train the student, our approach delivers higher efficiency, although there is small accuracy loss compared to directly distilling with private data.

\begin{algorithm}[tb]
\small
\caption{\small{Differentially private synthetic distillation.}}
\label{alg:DPSD}
\begin{algorithmic}[1] 
\REQUIRE Iteration number $T$, batch size $b$, learning rates $\gamma_s$ and $\gamma_g$, the teacher $\Phi_t$ with parameters $\bl{w}_t$, the student $\Phi_s$ with parameters $\bl{w}_s$ and the generator $\Phi_g$ with parameters $\bl{w}_g$.
\STATE Initialize $\bl{w}_s^{(0)}$ and $\bl{w}_g^{(0)}$ with Xavier, generate a batch of random vectors $\bl{Z}^{(0)}=\{\bl{z}^{(0)}_i\}_{i=1}^{b}$, and set $t=0$.
\FOR {$t < T$}
\STATE Generate a batch of synthetic samples $\{\bl{x}_i\}_{i=1}^{b}$ from $\bl{Z}^{(t)}$ with $\bl{x}_i=\Phi_g(\bl{z}^{(t)}_i;\bl{w}_g^{(t)})$
\STATE Perform differentially private annotation with Eq. \eqref{eq:noisyannotation} to get noisy labels $\{\hat{\bl{y}}_i\}_{i=1}^{b}$
\STATE Compute $\mc{L}_s$ with Eq. \eqref{eq:ls} and update the student with $\bl{w}_s^{(t+1)}=\bl{w}_s^{(t)}-\gamma_s\cdot{\partial\mc{L}_s}/{\partial \bl{w}_s^{(t)}}$
\STATE Compute $\mc{L}_g$ with Eq. \eqref{eq:generatorupdate}, update the generator with $\bl{w}_g^{(t+1)}=\bl{w}_g^{(t)}-\gamma_g \cdot{\partial (\mc{L}_s+\mc{L}_g)}/{\partial \bl{w}_g^{(t)}}$ and the random vectors with $\bl{z}_i^{(t+1)}=\bl{z}_i^{(t)}-\gamma_g \cdot{\partial (\mc{L}_s+\mc{L}_g)}/{\partial \bl{z}_i^{(t)}}$.  
\ENDFOR
\RETURN $\bl{w}_s^{(T)}$ and $\bl{w}_g^{(T)}$
\end{algorithmic}
\end{algorithm}
    
\subsection{Privacy Guarantee and Convergence Analysis}
\begin{figure}[!t]
\centering\includegraphics[width=1.0\linewidth]{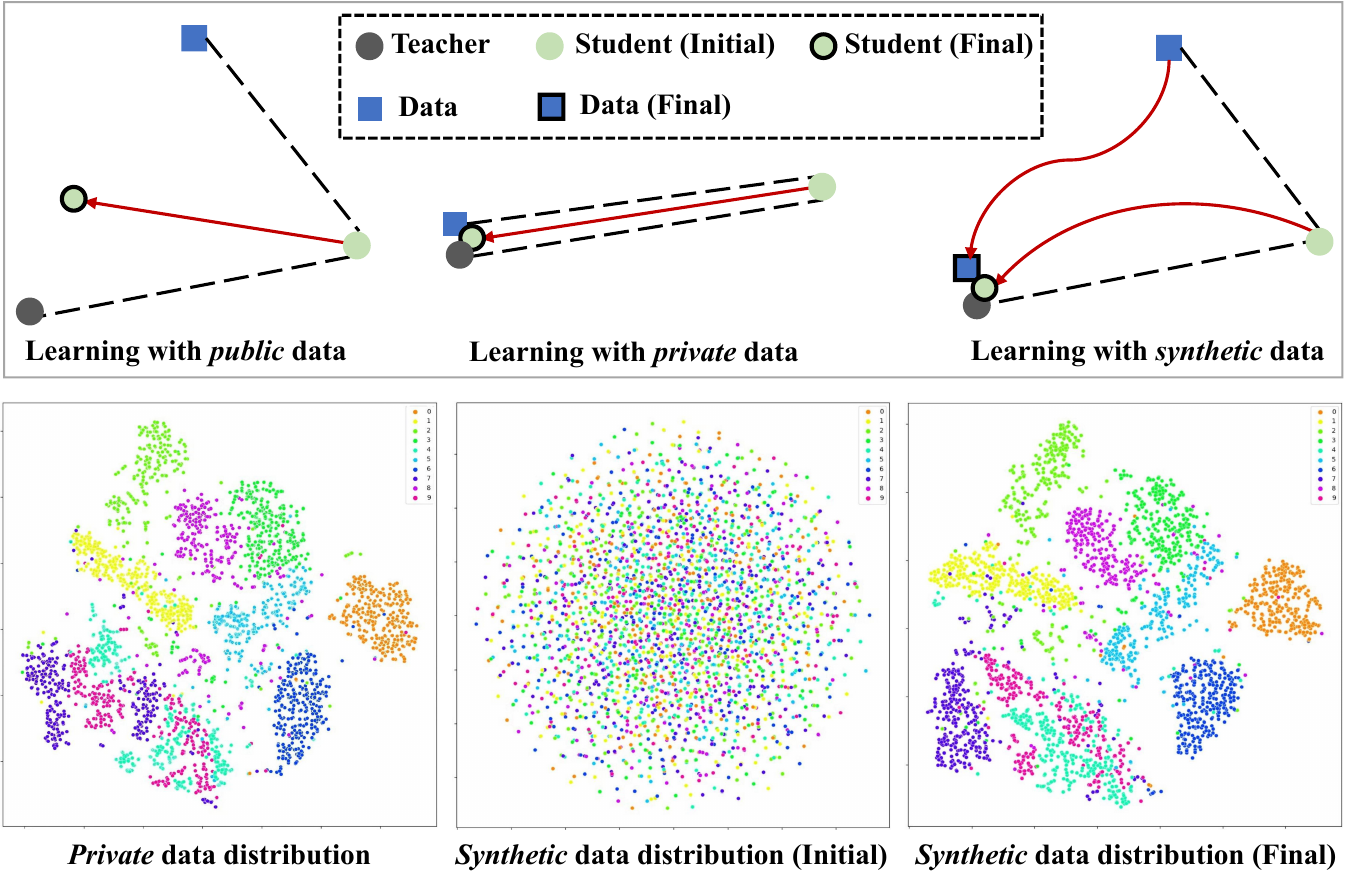}
\caption{Schematic student learning (Top). In existing approaches, learning with public data can avoid privacy issue but suffers from student convergence to the teacher, and learning with private data enables the student to converge to the teacher but leaks privacy. Our approach takes advantage of both sides by learning with synthetic data, where the final student converges to the teacher and the representation distribution of the generated synthetic data finally matches private data, \ie, achieving similar t-SNE representation visualization (Bottom).}\label{fig:subframework}
\end{figure}

Here, we demonstrate that our approach offers both privacy and convergence guarantees under the selections for data- and label-sensitive privacy protection, as depicted in Fig.~\ref{fig:subframework}.

\emph{1) Privacy Guarantee.}~For data-sensitive privacy protection, benefiting from the post-processing properties of differential privacy, it only requires a normalization process and noise addition onto the gradients of student outputs to derive new differentially private outputs (see Eq. \eqref{eq:dp}), which leads to three main advantages: i) higher accuracy with a smaller $\beta$ achieved by normalization due to relative scale maintaining among gradients, ii) lower privacy budget of teacher query due to the smaller $\beta$, and iii) privacy guarantee in theory (as seen in Theorem~\ref{th:dp-theorem} whose proof can be found in Append. B).
\begin{theorem}\label{th:dp-theorem}
\itshape{The data-sensitive privacy protection can guarantee $({2\beta^2cbTq}/{\sigma^2}+\log {((q-1)/q)}-\log{q\delta}/{(q-1)},\delta)$-DP for all $q>1$ and the failure probability $\delta\in(0,1)$.}
\end{theorem}

For label-sensitive privacy protection, only the labels from the teacher rather than the generated data need to be protected. Unlike common DP algorithms (\eg, DPSGD and PATE), it does not require the composition theorems to calculate the final privacy budget. Instead, it requires random labels once and then re-uses them throughout the training process. Thus, the amount of synthetic data and batch size have no impact on privacy, leading to a theoretical privacy guarantee in Theorem~\ref{th:theorem-label} (the proof is in Append. B).
\begin{theorem}\label{th:theorem-label}
\itshape{The label-sensitive privacy protection satisfies ($\varepsilon$,0)-LabelDP.}
\end{theorem}

\emph{2) Convergence Analysis.} Intuitively speaking, convergence of the model learning requires balancing synthetic data noise and student gradient updates. Thus, to achieve successful and stable training (\ie, convergence) of the student model under differential privacy constraints, we must carefully control and balance two factors: 1) the amount of noise added to protect privacy in the synthetic data or teacher outputs, and 2) the quality and reliability of the supervision signal received by the student, which directly affects its gradient updates. The detailed convergence analysis can be found in Append. C.

For data-sensitive privacy protection, the distillation loss $\mc{L}_s$ in Eq. \eqref{eq:ls} is optimized under differential privacy. We follow the standard assumptions in SGD~\cite{bottou2018optimization} with an additional assumption on gradient noise, and assume that $\mc{L}_s$ is $\kappa$-smoothness with a lower bound $\mc{L}_s^{(*)}$ and initial loss $\mc{L}_s^{(0)}$, described as: $\forall \bl{y},\bl{y}^{\prime}$, there is a constant $\kappa \geq 0$ such that $\mc{L}_s(\bl{y};\bl{w}_s)-\mc{L}_s(\bl{y}^{\prime};\bl{w}_s) \le \nabla \mc{L}_s(\bl{y};\bl{w}_s)^{\top}(\bl{y}-\bl{y}^{\prime}) + \frac{\kappa}{2}||\bl{y}-\bl{y}^{\prime}||^2$. The additional assumption is $(\bl{g}_r-\bl{g})\sim \mc{N}(0,\zeta^2)$, where $\zeta$ is Gaussian variance, $\bl{g}_r$ is the ideal gradients of $\mc{L}_s$ and $\bl{g}={\nabla \mc{L}_s}/{\nabla\bl{w}_s}$ is the gradient to be computed as an unbiased estimate of $\bl{g}_r$. Then according to~\cite{bu2022automatic}, we denote $\chi=\min\limits_{0 \leq t \leq T} \mathbb{E}(\|g_{r}^{(t)}\|)$ and have
\begin{equation}\nonumber
\chi \leq \mc{F}\left( \sqrt{\frac{2\left(\mc{L}_s^{(0)}-\mc{L}_s^{(*)}\right)+2T\kappa \gamma_s^2\beta^2(1+\sigma^2d)}{T\gamma_s \beta}} ; \zeta, h\right),
\end{equation}
where $d$ is a constant number. $\mc{F}(\cdot)$ results only from the normalization operation same as~\cite{bu2022automatic} and it won't affect the input monotonicity. With $\gamma_s \propto 1/\sqrt{T}$, the gradients will gradually tend to 0 as $T$ increases, implying the convergence. 

For label-sensitive privacy protection, we specifically consider the worst case wherein the gradient generated by the randomized response mechanism differs completely when the outputs and inputs vary. In this case, applying a randomized response mechanism to the label effectively becomes equivalent to its application on the gradient. Adhering to the same five assumptions in~\cite{bottou2018optimization}, we posit that (i) $||\nabla \Phi_s(\cdot;\bl{w}_s)-\nabla \Phi_s(\cdot;\bl{w}_s^{\prime})||_2\leq \kappa||\bl{w}_s-\bl{w}_s^{\prime}||_2$; (ii) $\Phi_s(\cdot;\bl{w}_s) \geq \Phi_s(\cdot;\bl{w}_s^{\prime}) + \nabla\Phi_s(\cdot;\bl{w}_s^{\prime})^{\top}(\bl{w}_s-\bl{w}_s^{\prime})+\frac{1}{2}c||\bl{w}_s-\bl{w}_s^{\prime}||_2^2$; (iii) $\nabla \Phi_s(\cdot;\bl{w}_s)^{\top}\mathbb{E}_{\bl{x}}[\bl{g}(\bl{x};\bl{w}_s)]\geq \mu||\nabla\Phi_s(\bl{x};\bl{w}_s)||_2^2$; (iv) $||\mathbb{E}_{\bl{x}}[\bl{g}(\bl{x};\bl{w}_s)]||_2\leq \mu_g||\nabla\Phi_s(\bl{x};\bl{w}_s)||_2$; and (v) $\mathbb{V}_{\bl{x}}[\bl{g}(\bl{x};\bl{w}_s)]\leq \mu_c + \mu_v||\nabla\Phi_s(\bl{x};\bl{w}_s)||_2^2$, where $\nabla\Phi_s(\bl{x};\bl{w}_s)$ is the true gradient, $\bl{g}(\bl{x};\bl{w}_s)$ is the computed gradient, $\mathbb{E}[\cdot]$ is for mean calculation, $\mathbb{V}[\cdot]$ is for variance calculation and $\kappa,c,\mu,\mu_g,\mu_v,\mu_c$ are non-negative constants. Then we get
\begin{equation}\nonumber
\begin{aligned}
&\mathbb{E}[\Phi_s(\cdot;\bl{w}^{(t+1)}_s)-\Phi_s(\cdot;\bl{w}_s^{(*)})]+\frac{\gamma_s^2\kappa \mu_c}{4\tau c}\\ \leq &(2\tau c+1)(\mathbb{E}[\Phi_s(\cdot;\bl{w}_s^{(t)})-\Phi_s(\cdot;\bl{w}_s^{(*)})]+\frac{\gamma_s^2\kappa \mu_c}{4\tau c}),
\end{aligned}
\end{equation}
where $\tau = -\gamma_s(\frac{2e^{\varepsilon}}{e^{\varepsilon}+k-1}-1)\mu+\frac{1}{2}\gamma_s^2\kappa(\mu_g^2+\mu_v)$. When we guarantee that $\tau < 0$, the algorithm will converge and the error from the minimum $\Phi_s(\cdot;\bl{w}_s^{(*)})$ is $-\frac{\gamma_s^2\kappa \mu_c}{4\tau c}$. 

\subsection{Discussion}
\emph{1) Extension to Federated Learning Setup.} In this setup, our approach transcribes distributed teachers into a student like~\cite{zhang2022dense}, where $m$ teachers $\{\Phi_t^j\}_{j=1}^{m}$ are independently trained on different clients. On the server, the generator generates a batch of synthetic data $\{\bl{x}_i\}_{i=1}^b$ using Alg.~\ref{alg:DPSD} that are used to train the global student, yielding predictions $\{\Phi_s(\bl{x}_i;\bl{w}_s)\}_{i=1}^b$. Both predictions and synthetic data are transmitted to each client, where the synthetic data are processed by each teacher to obtain differentially private labels. Finally, these labels are transmitted back to the server, facilitating the student update and generator update. The federated learning scenario achieves differential privacy via gradient averaging from multiple teachers. Our approach transmits synthetic data rather than the teachers, guaranteeing privacy even with an untrusted server. The algorithm can be found in Append.~\ref{appendix:fed}.

\emph{2) Relationships to Other Approaches.}~Our approach can be seen as the fusion of data-free knowledge distillation~\cite{chen2019data,fang2022up} and generative adversarial network (GAN) under differential privacy~\cite{xie2018differentially,jordon2018pate,chen2020gs}. It performs model transcription with the synthetic data from a trainable generator, and is very different from most existing approaches that access to private or public data. It utilizes a gradient-sanitized technique to reduce the noise addition like GS-WGAN~\cite{chen2020gs} and DataLens~\cite{wang2021ccs} but in the distillation process instead of adversarial training, which provides a better trade-off between privacy and performance. Different from the approaches that also use synthetic data for distillation~\cite{ge2023tip,liu2023model}, our unified framework provides switchable data-sensitive or label-sensitive privacy protection and an end-to-end competitive-cooperative learning by top-$k$ dimensionality reduction as well as taking the student prediction as prior into the differentially private annotation, which facilitates knowledge transfer and learning efficiency, \eg, only 33 minutes with 20 iterations to transcribe a ResNet34 into a three-layer CNN using three GeForce RTX 3090 GPUs.

\section{Experiments}
To verify the effectiveness of our DPSD approach, we conduct experiments on 8 datasets and comparisons with 26 state-of-the-arts. These baselines consist of three categories, including 17 data-sensitive privacy protection approaches, 4 label-sensitive privacy protection approaches and 5 approaches under federated learning setup. Among them, most approaches involve original \emph{private} data in their learning, and we keep their settings for the comparisons, while our DPSD uses \emph{synthetic} data generated by the trainable generator in the learning. We evaluate the test accuracy of all models under different privacy, \eg, transcribing pretrained models into their privacy-preserving counterparts and then evaluating their performance. To make the comparisons fair, our experiments use the same settings as these baselines and take the results from their original papers or by executing the official source codes.

\subsection{Experimental Setup}
\myPara{Datasets.} The datasets for evaluation include 5 popular benchmarks (MNIST~\cite{lecun1998gradient}, Fashion-MNIST (FMNIST)~\cite{fashionMnist}, CIFAR10~\cite{cifar10}, CIFAR100~\cite{cifar10} and ImageNet~\cite{deng2009imagenet}) for general image classification, and 3 datasets (CelebA~\cite{liu2015celeba}, MedMNIST~\cite{medmnistv2} and COVIDx~\cite{Wang2020covidx}) for specific image recognition. Similar to \cite{liu2023model}, we use two derivative datasets from CelebA, named CelebA-H and CelebA-G, designed for hair color classification (\eg, black, blonde or brown) and facial gender classification, respectively. COVIDx is a chest X-ray image dataset containing eight lesion categories for the detection of COVID-19 cases.

\myPara{Pretrained models.} To verify the flexibility of our DPSD approach over various network architectures, we study the transcription performance on 9 popular teacher architectures (AlexNet~\cite{krizhevsky2012imagenet}, VGGNet~\cite{vgg2015iclr}, ResNet~\cite{resnet2016cvpr}, WideResNet~\cite{zagoruyko2016wide}, DenseNet~\cite{huang2016deep}, MobileNet~\cite{howard2017mobilenets}, ShuffleNet~\cite{zhang2018shufflenet}, GoogleNet~\cite{szegedy2015going} and vision transformer (ViT)~\cite{dosovitskiy2020image}) and pretrain various models for evaluation. Specifically, we utilize a 19-layer VGGNet variant with batch normalization for all datasets, a 50-layer ResNet for ImageNet and 34-layer ResNet for other datasets, a 161-layer DenseNet with a growth rate of 24 for all datasets, and a ViT described in~\cite{dosovitskiy2020image} for all datasets. 
For the student model, we deploy a 34-layer ResNet for ImageNet and adopt the same network configuration as DataLens~\cite{wang2021ccs} for other datasets.

\myPara{Implementation details.} We set noise scale as $\sigma=100$. For data-sensitive privacy protection, we set the normalization bound as $\beta=10^{-3}$ when $\varepsilon \le 1$ and $\beta=5\times10^{-3}$ when $1 < \varepsilon \le 10$ for datasets except ImageNet and set $\beta=10^{-4}$ for ImageNet. In training, we set the batch size $b=256$, the failure probability $\delta=10^{-5}$ and the number of clients $m=5$ in the federated learning setup. For top-$k$ selection, we set $k=3$ for MNIST, FMNIST, CIFAR10 and MedMNIST, $k=50$ for CIFAR100, $k=2$ for CelebA-G, $k=3$ for CelebA-H and CoVIDx, and $k=300$ for ImageNet. The learning rates for student and generator update are set to $\gamma_s=0.1$ and $\gamma_g=0.01$, respectively. The total iteration number is set to $T=200$. We use ResNet50 as teacher and ResNet34 as student for ImageNet, and ResNet34 as teacher and the same structure in~\cite{wang2021ccs,Wang2020covidx,liu2023model} as student for fair comparisons. All experiments are implemented in Pytorch platform with Adam optimizer for training on three GeForce RTX 3090 GPUs.

\begin{figure*}[t]
\centering\includegraphics[width=1.0\linewidth]{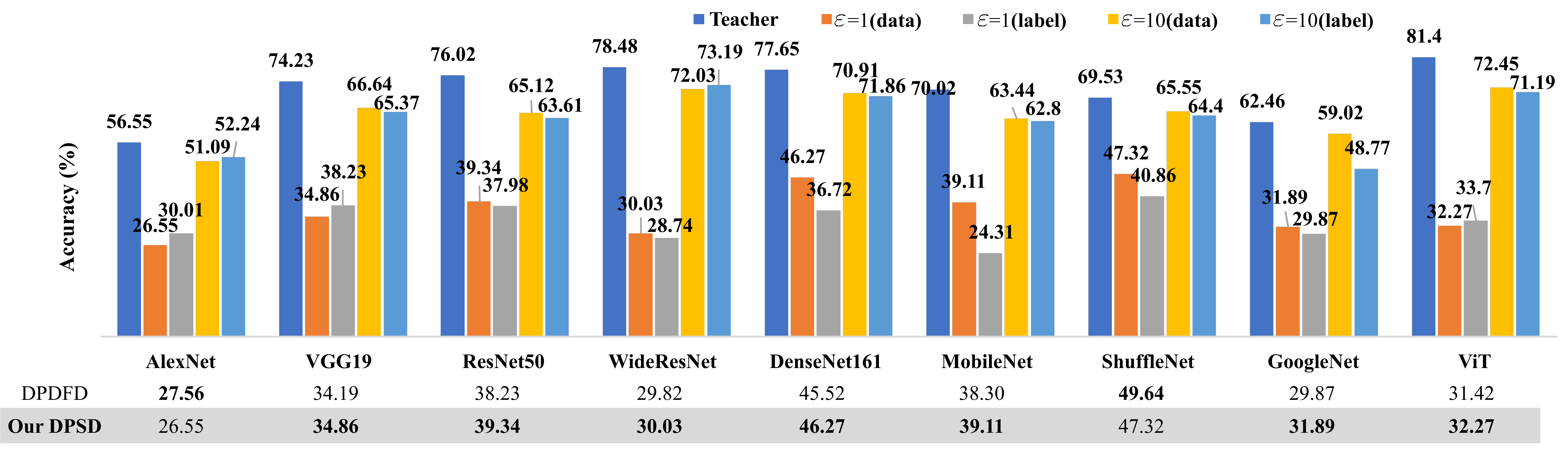}
\caption{\textbf{Evaluation on model transcription}. We show the test accuracy of transcribed ResNet34 students on ImageNet with different pretrained deep learning models as teachers under a low failure probability of $\delta=10^{-5}$ and different privacy budget $\varepsilon$.  We also provide a comparison to the recent DPDFD~\cite{liu2023model} under $\varepsilon=1$, showing better accuracy achieved by our DPSD.}
\label{fig:networks}
\end{figure*}

\begin{table*}[t]
\setlength{\tabcolsep}{4.5pt}%
\renewcommand\arraystretch{1.25}
\caption{Test accuracy (\%) comparisons with 12 explicit data-sensitive privacy protection approaches under a low failure probability $\delta=10^{-5}$ and two privacy budgets $\varepsilon$ (1 and 10).}\label{tab:dp(1_and_10)}
\small
\begin{center}\begin{threeparttable}\begin{tabular}{lccccccccccccc}
\hline
\multirow{2}{*}{Approach} &\multirow{2}{*}{Involved data} &&\multicolumn{2}{c}{{MNIST}} &&\multicolumn{2}{c}{{FMNIST}}&&\multicolumn{2}{c}{{CelebA-G}}&&\multicolumn{2}{c}{{CelebA-H}}\cr
\cline{4-5} \cline{7-8} \cline{10-11} \cline{13-14}&&& $\varepsilon$=1 & $\varepsilon$=10 && $\varepsilon$=1 & $\varepsilon$=10 && $\varepsilon$=1 & $\varepsilon$=10 && $\varepsilon$=1 & $\varepsilon$=10\cr
\hline
\multirow{1}{*}{\textcolor{gray}{Teacher}} &\multirow{1}{*}{\textcolor{gray}{Private}} &&\multicolumn{2}{c}{\textcolor{gray}{99.21}} &&\multicolumn{2}{c}{\textcolor{gray}{91.02}}&&\multicolumn{2}{c}{\textcolor{gray}{93.53}}&&\multicolumn{2}{c}{\textcolor{gray}{88.68}}\cr
\hline
{DP-GAN~\cite{xie2018differentially}} (arXiv'18) & Private \& Synthetic && 40.36 & 80.11 && 10.53 & 60.98 && 53.30 & 52.11 && 34.47 & 39.20\cr
{PATE-GAN~\cite{jordon2018pate}} (ICLR'19) & Private \& Synthetic && 41.68 & 66.67 && 42.22 & 62.18 && 60.68 & 65.35 && 37.89 & 39.00\cr
{GS-WGAN~\cite{chen2020gs}} (NeurIPS'20) & Private \& Synthetic && 14.32 & 80.75 && 16.61 & 65.79 && 59.01 & 61.36 && 42.03 & 52.25\cr
{DP-MERF~\cite{harder2020differentially}} (AISTATS'21) & Private \& Synthetic && 63.67 & 67.38 && 58.62 & 61.62 && 59.36 & 60.82 && 44.13 & 44.89\cr
{P3GM~\cite{takagi2021p3gm}} (ICDE'21) & Private \& Synthetic && 73.69 & 79.81 && 72.23 & 74.80 && 56.73 & 58.84 && 45.32 & 48.58\cr
{G-PATE~\cite{long2021g}} (NeurIPS'21) & Private \& Synthetic && 58.10 & 80.92 && 55.67 & 69.34 && 67.02 & 68.97 && 49.85 & 62.17\cr
{DataLens~\cite{wang2021ccs}} (CCS'21) & Private \& Synthetic && 71.23 & 80.66 && 64.78 & 70.61 && 70.58 & 72.87 && 60.61 & 62.24\cr
{DPSH~\cite{cao2021don}} (NeurIPS'21) & Private \& Synthetic && -- & 83.20 && -- & 71.10 && -- & 76.30 && -- & --\cr
{DPGEN~\cite{chen2022dpgen}} (CVPR'22) & Private \& Synthetic && 90.46 & 93.57 && 82.83 & 87.84 && 69.99 & 88.35 && 66.14 & 81.47\cr
{PSG~\cite{chen2022privateset}} (NeurIPS'22) & Private \& Synthetic && 80.90 & 95.60 && 70.20 & 77.70 && -- & -- && -- & --\cr
{DGD~\cite{ge2023tip}} (TIP'23) & Private \& Synthetic && 88.20 & 97.40 && 64.90 & 73.60 && -- & -- && -- & --\cr
{DPDFD~\cite{liu2023model}} (IJCAI'23) & Synthetic && 95.12 & 97.51 && 83.86 & 89.88 && 72.37 & 89.92 && 78.39 & 82.35\cr
\rowcolor{lightgray!30}
\textbf{Our DPSD} & Synthetic && \textbf{96.03} & \textbf{97.85} && \textbf{83.97} & \textbf{90.08} && \textbf{73.74} & \textbf{91.23} && \textbf{79.21} & \textbf{83.02}\cr
\hline
\end{tabular}\end{threeparttable}\end{center}
\end{table*}

\subsection{Evaluation on Model Transcription}
Our DPSD approach can transcribe various models into privacy-preserving ones. To verify that, we take 9 popular deep learning models pretrained on ImageNet from Pytorch model zoo as teachers and check the accuracy of transcribed students under two privacy selections as well as different privacy budgets. The networks of these models cover classic convolutional neural networks, light-weight networks and Transformer. We use ResNet34 for the student network and report the results in Fig.~\ref{fig:networks}. As expected, we find that the accuracy of all transcribed students drops sharply under a small privacy budget (\eg, $\varepsilon=1$) and gently under a large privacy budget (\eg, $\varepsilon=10$), which inherits the characteristic of knowledge distillation where seriously noisy labels will reduce the knowledge transfer. We also find that under the small privacy budget of $\varepsilon=1$ the student transcribed from ShuffleNet delivers the highest accuracy of 47.32\% while the teacher accuracy is 11.87\% lower than the highest ViT (\eg, 81.40\%), implying that a better teacher is more likely to facilitate knowledge transfer under low privacy budget. Moreover, as $\varepsilon$ increases to 10, a transcribed student can learn more clear knowledge from teacher and its accuracy is mainly dominated by its teacher, suggesting that suitable student network can accept the teacher knowledge and facilitate generator learning. We also find that under $\varepsilon=5$ the student accuracy is positively correlated with the teacher accuracy, which proves that our DPSD is not only scalable but also effective to various popular networks. To further show the effectiveness of our approach, we compare the transcription performance with the recent DPDFD approach. Fig.~\ref{fig:networks} shows that our DPSD delivers higher test accuracy than DPDFD in most cases.

\subsection{Evaluation on Data-Sensitive Privacy Protection}
The 17 data-sensitive privacy protection approaches including 12 explicit approaches trained with generative data (DP-GAN~\cite{xie2018differentially}, GS-WGAN~\cite{chen2020gs}, PATE-GAN~\cite{jordon2018pate}, DP-MERF~\cite{harder2020differentially}, P3GM~\cite{takagi2021p3gm}, DataLens~\cite{wang2021ccs}, G-PATE~\cite{long2021g}, DPSH~\cite{cao2021don}, DPGEN~\cite{chen2022dpgen}, PSG~\cite{chen2022privateset},  DGD~\cite{ge2023tip} and DPDFD~\cite{liu2023model}) and 5 implicit approaches trained with differentially private learning (DPSGD~\cite{abadi2016deep}, TSADP~\cite{papernot2021tempered}, TOPAGG~\cite{wang2021ccs}, GM-DP~\cite{mcmahan2018general} and Spectral-DP~\cite{feng2023sp}).

\begin{table}[t]
\setlength{\tabcolsep}{3.5pt}\renewcommand\arraystretch{1.25}
\caption{Test accuracy (\%) comparisons on two medical datasets under data-sensitive privacy protection.}\label{tab:dp(med_and_covidx)}\begin{center}
\small
\begin{threeparttable}\begin{tabular}{lccccc}
\hline
\multirow{2}{*}{Approach} &\multicolumn{2}{c}{{MedMNIST}}&&\multicolumn{2}{c}{{COVIDx}}\cr
\cline{2-3} \cline{5-6}
& $\varepsilon$=1 & $\varepsilon$=10 && $\varepsilon$=1 & $\varepsilon$=10\cr
\hline
{DataLens~\cite{wang2021ccs} (CCS'21)} & 53.52 & 68.90 && 48.61 & 55.87\cr
{DPDFD~\cite{liu2023model} (IJCAI'23)} & 80.12 & 85.89 && 73.61 & 81.69\cr
\rowcolor{lightgray!30}
\textbf{Our DPSD} & \textbf{81.25} & \textbf{86.01} && \textbf{74.11} & \textbf{81.73}\cr
\hline
\end{tabular}\end{threeparttable}\end{center}
\end{table}

\emph{1) Comparisons with 12 Explicit Approaches.} We evaluate all approaches under a low failure probability of $\delta=10^{-5}$ and report the results in Tab.~\ref{tab:dp(1_and_10)} and~\ref{tab:dp(med_and_covidx)}. We find several meaningful conclusions. First, our DPSD consistently delivers higher test accuracy than other approaches under various privacy budgets, even training on synthetic data. Specifically, under $\varepsilon=1$, it still achieves an accuracy of 96.03\% on MNIST and 83.97\% on FMNIST, which remarkably reduces the accuracy drop against the teacher model by 3.18\% and 7.05\% respectively. Second, on two high dimensional CelebA datasets whose dimensionality are about 16 times larger than MNIST, most approaches suffer from big accuracy drop even under a high privacy budget of 10 and our approach can remain good accuracy under $\varepsilon=10$, \eg, the accuracy drop by 2.30\% on CelebA-G and 5.66\% on CelebA-H respectively. Third, on two realistic medical datasets shown in Tab.~\ref{tab:dp(med_and_covidx)}, DPSD outperforms DataLens and DPDFD, implying its best privacy-preserving ability with minimal accuracy drop in piratical applications.

\emph{2) Comparisons with 5 Implicit Approaches.} As shown in Tab.~\ref{tab:implicit}, we can find that our DPSD achieves the highest accuracy of 96.03\% on MNIST and 86.59\% on CIFAR10 under the lowest privacy budget of 1.0 and 2.0, respectively. We suspect the main reason lies in the usage of top-$k$ selection as well as normalization instead of traditional clipping in data protection or randomized response in label protection, enabling the obtained differentially private gradients preserve more information due to relative scale between gradients.

\begin{table}[t]
\setlength{\tabcolsep}{3.0pt}%
\renewcommand\arraystretch{1.3}
\caption{Test accuracy (\%) comparisons with 5 implicit approaches under data-sensitive privacy protection.}\label{tab:implicit}
\small
\begin{center}\begin{threeparttable} \begin{tabular}{lcc|cc}
\hline
{Approach} & $\varepsilon$ & {MNIST} & $\varepsilon$ & {CIFAR10}\cr
\hline
{DPSGD~\cite{abadi2016deep} (CCS'16)} & 2.0 & 95.00 & 2.0 & 66.23\cr
{GM-DP~\cite{mcmahan2018general} (arXiv'18)} & 1.0 & 95.08 & 2.0 & 85.97\cr
    {TSADP~\cite{papernot2021tempered} (AAAI'21)} & 1.0 & 79.91 & 7.5 & 66.20\cr
    {TOPAGG~\cite{wang2021ccs} (CCS'21)}& 1.0 & 94.65 & 2.0 & 85.18\cr
    {Spectral-DP~\cite{feng2023sp} (SP'23)}  & 2.0 & 95.77 & 3.0 & 69.51\cr
\rowcolor{lightgray!30}
\textbf{Our DPSD}& 1.0 & \textbf{96.03} & 2.0 & \textbf{86.59}\cr
\hline
\end{tabular}\end{threeparttable}\end{center}
\end{table}

\subsection{Evaluation on Label-Sensitive Privacy Protection}
Our approach can flexibly switch to label-sensitive privacy protection. To demonstrate that, we compare with 4 approaches with LabelDP protection (LP-MST~\cite{ghazi2021nips}, ALIBI~\cite{malek2021nips}, ClusterRR~\cite{esfandiari2022aistats} and Protocol~\cite{yuan2021nipsws}) on MNIST, FMNIST, CelebA-H and CelebA-G under the same $\varepsilon$. The results are shown in Tab.~\ref{tab:LabelDP}. We can find that our approach performs optimally for all four datasets and for different $\varepsilon$. In particular, it achieves an accuracy of 74.67\% when $\varepsilon=8$ on CelebA-G, surpassing many approaches that use the original private data for direct distillation. The effectiveness of our approach is further demonstrated by the fact that it delivers better performance than other four approaches trained directly using raw data when there is no restriction on the amount of generated synthetic data.

\begin{table}[t]
\setlength{\tabcolsep}{2.5pt}\renewcommand\arraystretch{1.25}
\caption{Test accuracy (\%) comparisons with 4 approaches under label-sensitive privacy protection.}\label{tab:LabelDP}
\small
\begin{center}
\begin{threeparttable}\begin{tabular}{lccccccccc}
\hline
\multirow{2}{*}{Approach} &{MNIST} &&{FMNIST}&&\multicolumn{2}{c}{{CelebA-H}}&&\multicolumn{2}{c}{{CelebA-G}}\cr
\cline{2-2} \cline{4-4} \cline{6-7} \cline{9-10}
& $\varepsilon$=1 && $\varepsilon$=1 && $\varepsilon$=1 & $\varepsilon$=2 && $\varepsilon$=3 & $\varepsilon$=8\cr
\hline
{LP-2ST~\cite{ghazi2021nips}} & 95.82 && 83.26 && 63.67 & 86.05 && 28.74 & 74.10\cr
{ALIBI~\cite{malek2021nips}} & -- && -- && 84.20 & -- && 55.00 & 74.40\cr
{Protocol~\cite{yuan2021nipsws}} & -- && -- && -- & 81.84 && -- & --\cr
{ClusterRR~\cite{esfandiari2022aistats}} & 90.00 && 88.00 && 68.57 & -- && -- & --\cr
\rowcolor{lightgray!30}
\textbf{Our DPSD} & \textbf{97.62} && \textbf{89.17}  && \textbf{87.96} & \textbf{88.12} && \textbf{58.61} & \textbf{74.67}\cr
\hline
\end{tabular}\end{threeparttable}\end{center}
\end{table}

\subsection{Evaluation under Federated Learning Setup}
To verify the effectiveness under federated learning setup, we compare our variant FedDPSD with baseline FedAVG~\cite{mcmahan2017communication} as well as four distillation-based approaches (FedDF~\cite{lin2020ensemble}, FedDAFL~\cite{chen2019data}, FedADI~\cite{yin2020dreaming} and DENSE~\cite{zhang2022dense}). To simulate real-world applications, we use the same setup as in~\cite{zhang2022dense} by using Dirichlet distribution with a scaling parameter of 0.5 to generate non-IID data partitions for five disjoint clients. Each client holds its dataset private and the serve only has access to client models. Our federated variant quadratically reduces the impact of differential privacy noise by averaging a batch of gradients due to multiple teachers, and sets the privacy budget as 10. The results are reported in Tab.~\ref{tab:fed}. It can be found that our FedDPSD achieves superior performance. In particular, it outperforms other approaches on high dimensional dataset ImageNet. This shows the scalability of our approach.

\begin{table}[t]
\setlength{\tabcolsep}{2.0pt}%
\renewcommand\arraystretch{1.3}
\caption{Test accuracy (\%) comparisons with 5 approaches under federated learning setup and five non-IID data partitions with Dirichlet distribution.}\label{tab:fed}
\small
\begin{center}
\begin{threeparttable}
\begin{tabular}{lcccc}
\hline
{Approach} &{MNIST} &{FMNIST} &{CIFAR10} &{ImageNet}\cr
\hline
{FedAVG~\cite{mcmahan2017communication} (AISTATS'17)} & 90.55 & 83.72 & 43.67 & 11.89\cr
{FedDAFL~\cite{chen2019data} (ICCV'19)}& 93.01 & 84.02 & 58.59 & 28.22\cr
{FedDF~\cite{lin2020ensemble} (NeurIPS'20)} & 92.18 & 84.67 & 53.56 & 27.43\cr
{FedADI~\cite{yin2020dreaming} (CVPR'20)} & 93.49 & 84.19 & 59.34 & 30.21\cr
{DENSE~\cite{zhang2022dense} (NeurIPS'22)}  & \textbf{95.82} & 85.94 & 62.19 & 32.34\cr
\rowcolor{lightgray!30}
\textbf{Our FedDPSD}& {95.76} & \textbf{88.64} & \textbf{71.12} & \textbf{39.13}\cr
\hline
\end{tabular}\end{threeparttable}\end{center}
\end{table}

\begin{figure*}[ht]
	\centering
    \includegraphics[width=1.0\linewidth]{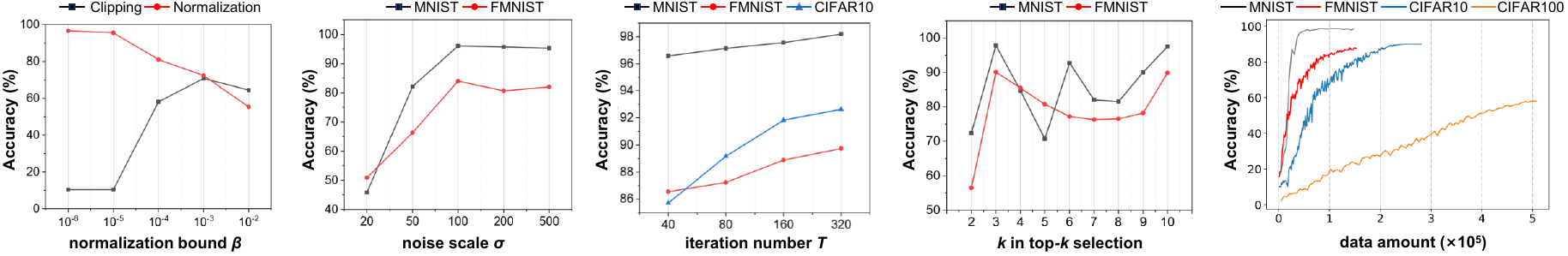}
\caption{The effect of the key operations and parameter settings in differentially private annotation.}
	\label{fig:norm_sigma_stage_data}
\end{figure*}

\subsection{Ablation Studies}
After the promising performance is achieved, we further analyze the impact of each component in our approach, including differentially private annotation, teacher network, the choice of discriminator for generator training, the loss terms for generator training, and the distillation approach. 

\emph{1) Differentially Private Annotation.} This process is very critical to balance model accuracy and privacy protection. We train models under label-sensitive privacy protection to study the effect of its key operations and parameter settings and report results in Fig.~\ref{fig:norm_sigma_stage_data}. First, normalization operation can limit the gradients' sensitivity and is better than traditional clipping under small bound $\beta$. Its superiority against clipping will gradually decrease as $\beta$ increases, since it retains the relative scale information of the gradients while clipping retains the absolute one, suggesting a certain normalization bound (\eg, $\beta=10^{-5}$). Second, the setting of noise scale $\sigma$ is important in privacy protection. Fig.~\ref{fig:norm_sigma_stage_data} shows that the model accuracy gets better as $\sigma$ increases from 20 to 100 due to low privacy budget led by large noise scale, but slightly worse as $\sigma$ continues to increase since the gradients will be broken with a large noise scale. Thus, a trade-off should be made by setting a certain value (\eg, $\sigma=100$). Third, iteration number $T$ is related to synthetic data, thus increasing it from 40 to 320 can boost model accuracy and the effect is more obvious for a harder CIFAR100 dataset. The results are as expected since we use student prediction as prior to get more accurate prediction, resulting in clearer annotations on synthetic data to improve student performance. Fourth,  top-$k$ selection is important. As $k$ increases, the performance initially improves, then declines, and subsequently rises again. This demonstrates the balance between privacy and performance influenced by top-$k$ selection. Fifth, the amount of generated synthetic data is also important. The accuracy converges when data amount is about 50,000@MNIST, 120,000@FMNIST, 220,000@CIFAR10 and 500,000@CIFAR100, respectively, suggesting that a harder private dataset usually requires generating more synthetic data to learn its representation distribution.

\emph{2) Effect of Teacher.} The teacher pretrained from private data can provide knowledge to guide model transcription without direct data access, thus student learning needs sufficient knowledge distillation and effective knowledge transfer under differential privacy. To analyze the effect of the teacher in student learning, we conduct experimental comparison to two directly trained CNN models same as~\cite{chen2022privateset} on MNIST under $\varepsilon=1$, which achieve test accuracy of 97.65\% and 97.78\% under data-sensitive and label-sensitive privacy protection, respectively. As expected, the model trained directly on private data delivers better accuracy than its corresponding transcribed student. However, the accuracy achieved by our students only gets a small drop, \eg, 1.52\% and 0.16\% under data-sensitive and label-sensitive privacy protection respectively, while is higher than other approaches (see Tab. \ref{tab:dp(1_and_10)}, \ref{tab:implicit} and \ref{tab:LabelDP}) even with the same network structures. This implies that our approach can effectively distill teacher knowledge for model transcription with high accuracy under privacy guarantee.

\emph{3) Generator Training.}  We investigate the effect of discriminator and loss terms in generator training. Our approach can choose the pretrained teacher as a fixed discriminator for generator learning, leading to a variant termed as DPSD-T. In this case, the transcribed student still holds privacy-preserving ability but the learned generator does not have differential privacy guarantee. Here, by choosing teacher and student as discriminator, we conduct experimental comparisons on MNIST and FMNIST. It is found that the transcribed student by DPSD achieves a slightly lower accuracy than the one by DPSD-T, \eg, 95.85\% vs. 97.93\% on MNIST and 90.08\% vs. 91.24\% on FMNIST. These results show the effectiveness as well as utility of our default DPSD in protecting the privacy of both student and generator. 

\begin{table}[t]
\setlength{\tabcolsep}{7.6pt}%
\renewcommand\arraystretch{1.3}
\caption{Test accuracy (\%) of students trained on synthetic data under $\varepsilon=10$ and different losses in $\mc{L}_g$ and. CE: Cross entropy loss; IE: Information entropy loss; NORM: $l_2$-normalization loss. D and L mean data-sensitive and label-sensitive privacy protection, respectively.}\label{tab:loss}
\small
\begin{center}
\begin{threeparttable}
\begin{tabular}{lccccc}
	\hline
	\multirow{2}{*}{{Dataset}}& \multirow{2}{*}{{CE}} & \multirow{2}{*}{{IE}} & \multirow{2}{*}{{NORM}} & \multicolumn{2}{c}{{Accuracy}} \cr
    \cline{5-6}
    & & & & D & L \cr
	\hline
				\multirow{4}{*}{{MNIST}} & $\surd$ & $\surd$ & $\surd$ & \textbf{97.85} & \textbf{98.19}\cr
				&  & $\surd$ & $\surd$ & 96.02 & 96.55\cr
				& $\surd$ &  & $\surd$ & 66.37 & 94.32\cr
				& $\surd$ & $\surd$ &  & 86.39 & 88.01\cr
				\hline
				\multirow{4}{*}{{FMNIST}} & $\surd$ & $\surd$ & $\surd$ & \textbf{90.08} & \textbf{89.74}\cr
				&  & $\surd$ & $\surd$ & 79.46 & 88.04 \cr
				& $\surd$ &  & $\surd$ & 62.37 & 87.94 \cr
				& $\surd$ & $\surd$ &  & 83.00 & 81.22 \cr
				\hline
			\end{tabular}
		\end{threeparttable}
	\end{center}
\end{table}

\begin{figure}[ht]
\centering
\includegraphics[width=1.0\linewidth]{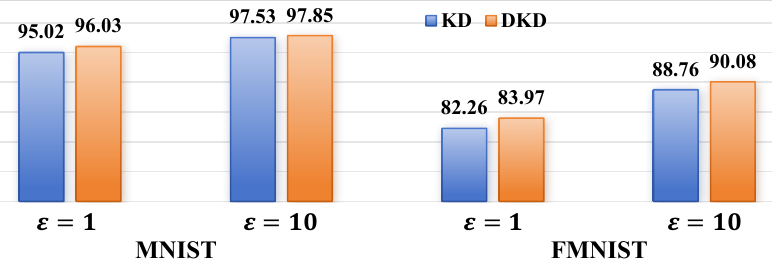}
\caption{Test accuracy (\%) of students with classic knowledge distillation (KD) and decoupled knowledge distillation (DKD).}
\label{fig:effect-distillation}
\end{figure}
 
\emph{4) Effect of Loss Terms.} The effect of three loss terms in Eq.~\eqref{eq:generatorupdate} is reported in Tab.~\ref{tab:loss}. For data-sensitive privacy protection, information entropy loss is the most important term to balance synthetic data classes, cross entropy loss contributes differently for different datasets (\eg, 1.0\% improvement on MNIST and 13\% improvement on FMNIST), and normalization loss is less critical but still useful, contributing to an accuracy improvement of 2\%-3\%. For label-sensitive privacy protection, normalization loss has the greatest impact, followed by information entropy loss and cross entropy loss. We speculate that this may be related to the randomness of the generated synthetic data, which limits the representation distribution to make the generated data more usable, so it has a greater impact on student accuracy. In summary, all three losses have a positive effect on model performance.

\emph{5) Effect of Distillation.} To check distillation method, we apply classic knowledge distillation~\cite{hinton2015distilling} to train models on MNIST and FMNIST and compare them with the models trained with decoupled knowledge distillation. From the results in Fig.~\ref{fig:effect-distillation}, as expected, the models with decoupled knowledge distillation deliver higher accuracy, suggesting that better distillation method can facilitate model transcription.

\subsection{Visualization Analysis}
Beyond promising privacy guarantee in theory, we check the intuitive reason of privacy-preserving capacity via visualizing the generated synthetic and reconstruction-attacked images. 

\begin{figure*}[!htb]
\centering
\includegraphics[width=1.0\linewidth]{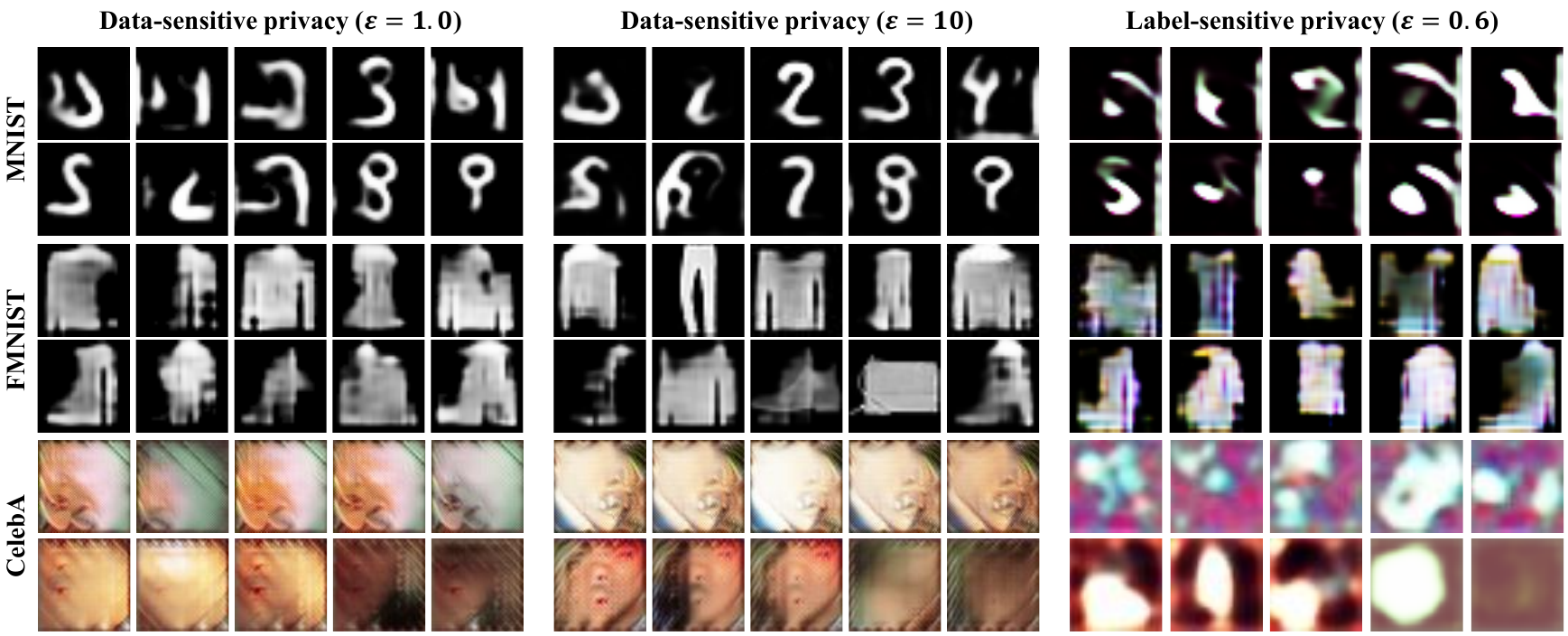}
\caption{Some synthetic images generated by the generators for three datasets under different privacy requirements.}
\label{fig:vis}
\end{figure*}

\emph{1) Visualization of Generated Synthetic Images.} Fig.~\ref{fig:vis} presents the synthetic images for MNIST, FMNIST and CelebA, respectively. In our approach, the learning goal is to train a generator to generate images with high utility in model training rather than high visual quality. Thus, these synthetic images usually have heavy noise but are still effective to train useful models. As expected, the visual quality of the synthetic images is higher under larger privacy budget. However, for the high-dimensional CelebA face dataset, the details of the synthetic images are still poor. We also visualize some synthetic examples under label-sensitive privacy protection. In this case, although we don't protect the images in theory, these synthetic images still have a lot of noise, probably because the data-free knowledge distillation framework is originally protective of the synthetic images. 

\begin{figure}[!htbp]
\centering
\includegraphics[width=1\columnwidth]{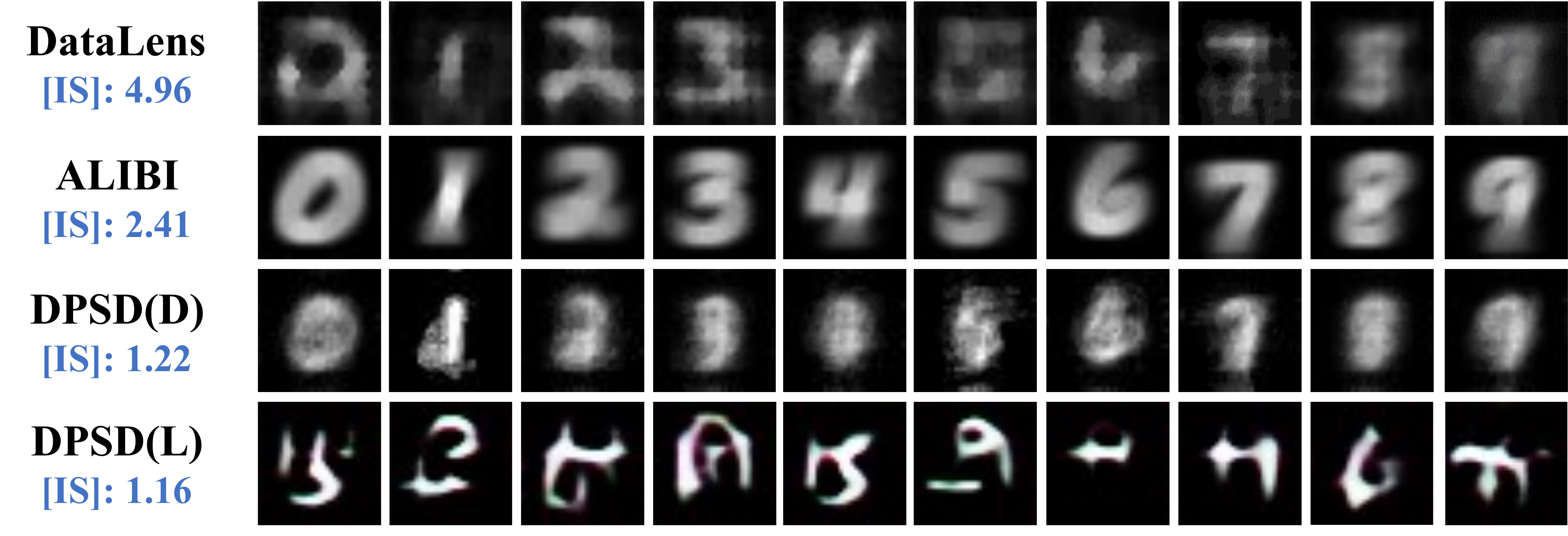}
\caption{Generative attack results over different privacy-preserving models on MNIST. We provide the average Inception score (IS) for evaluating the quality of generated images where [IS]=(IS-1)*$10^6$. Our models DPSD(D) and DPSD(L) deliver lower scores, implying the generated images are more difficult to be identified and have better privacy protection.}
\label{fig:attack-mf}
\end{figure}

\emph{2) Visualization of Attack Results.}~To further demonstrate the protection capability of our approach, we conduct generative inversion attack~\cite{zhang2020secret} experiments on the student models trained on MNIST. The privacy budget $\varepsilon$ of these models is set to 1. We compare our models DPSD(D) and DPSD(L) with two state-of-the-art approaches, DataLens under data-sensitive privacy protection and ALIBI under label-sensitive privacy protection, and show the generative attack results in Fig.~\ref{fig:attack-mf}. From the Inception scores as well as the attack-reconstructed images, we can find that attackers are more difficult to identify whether a particular image attacked from our models is in the original training data by inversion attack, implying better privacy-preserving ability. Although \cite{zhang2020secret} has stressed that differential privacy hardly works against the attack, we can find that even for experiments on the simplest MNIST dataset, our approach still can defend against the attack and protect the privacy of the private data.
 
\subsection{Further Discussion}
\emph{1) Limitation on Synthetic Data Generation.}
From Fig.~\ref{fig:vis}, we can find that the generated data, particularly for high-dimensional CelebA images, contain noise and artifacts. Although the objective of our approach is to generate privacy-preserving data, the generation quality of synthetic high-dimensional data still greatly affects the accuracy of privacy-preserving models, \eg, delivering a low accuracy on CelebA-H or CelebA-G in Tab.~\ref{tab:dp(1_and_10)} and on ImageNet in Fig.~\ref{fig:networks}. We suspect the main reasons come from data-free supervision from pretrained teacher as well as the use of GAN-based generators. A potential improvement is adopting more advanced  diffusion-based generators~\cite{liu2024eccv,esser2024icml,chen2025dc} in the framework.

\emph{2) Hyperparameter Sensitivity.}
In our approach, certain hyperparameters are used to control the process of differentially private annotation. For example, the normalization bound $\beta$ and noise scale $\sigma$ are used for data-sensitive privacy protection, and $k$ in top-$k$ selection is for label-sensitive privacy protection. Fig.~\ref{fig:effect-distillation} shows that the model accuracy is highly dependent on their careful tuning. The main reasons lie in: i) these parameters greatly affect the noise degree of labels by Eq.~\eqref{eq:dp} or Eq.~\eqref{eq:labelprivacy}, and ii) the student networks are directly learned on noisy labels. Thus, some feasible solutions including experiential parameter setting, and the use of label-noise learning algorithms~\cite{li2022cvpr,li2024tpami,zhang2024tpami} to improve model training under noisy labels.

\begin{figure}[!htbp]
\centering
\includegraphics[width=1\columnwidth]{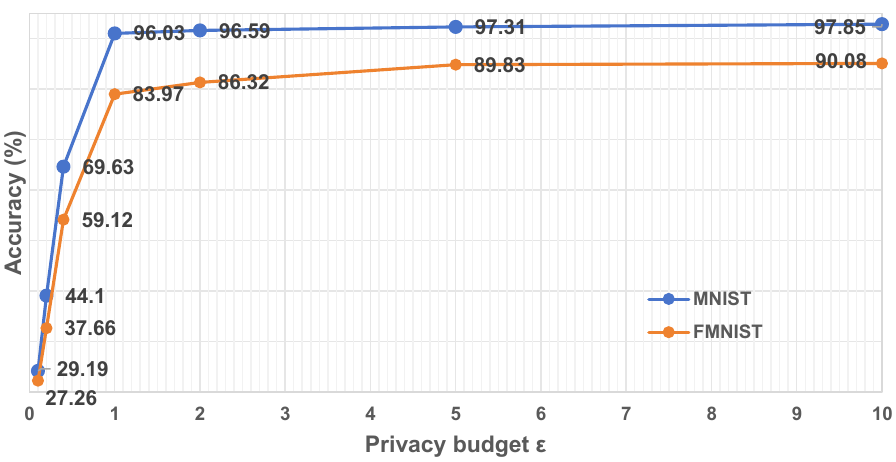}
\caption{Privacy-accuracy curve on MNIST and FMNIST under data-sensitive privacy protection. It shows serious accuracy degradation under extreme privacy constraints.}
\label{fig:privacy-accuracy}
\end{figure}

\emph{3) Privacy-Utility Tradeoff.}
To study that, we conduct an experiment on MNIST and FMNIST under data-sensitive privacy protection and extreme privacy constraints, and show the privacy-accuracy curve for $\epsilon\in[0.1,10]$ in Fig.~\ref{fig:privacy-accuracy}. We can see that the classification accuracy sharply decreases under a very low privacy budget, \eg, only 29.19\% on MNIST and 27.26\% on FMNIST under $\epsilon=0.1$. The main reason is obvious, which comes from large label noise caused by extreme low privacy budgets. One potential solution to improve student training under noisy labels is label-noise learning methods~\cite{li2022cvpr,li2024tpami,zhang2024tpami}. It also suggests the necessity of finding the range of values of which achieves a balance between utility and privacy~\cite{jayaraman2019uss}, \eg, setting a relatively low privacy budget of around 1 for an easier classification task on MNIST and a high privacy budget of around 5 on FMNIST.

\section{Conclusion}
Model privacy is critical in practical deployment. In this paper, we proposed a model transcription approach to convert pretrained teacher models to privacy-preserving student models without access to private data by differentially private synthetic distillation. The approach introduces a trainable generator into teacher-student learning paradigm and engages three players in a unified competitive-cooperative learning framework, which provides a flexible protection towards data or label privacy with theoretical guarantee. Extensive experiments have proven the effectiveness of the proposed approach. The future works include improving the approach with more advanced generators and expanding its application to more fields like large vision models. 

\myPara{Acknowledgements.}~This work was supported by grants from the Pioneer R\&D Program of Zhejiang Province (2024C01024).

\bibliographystyle{IEEEtran}
\bibliography{bibieee}

\begin{IEEEbiography}[{\includegraphics[width=1in,height=1.25in, clip,keepaspectratio]{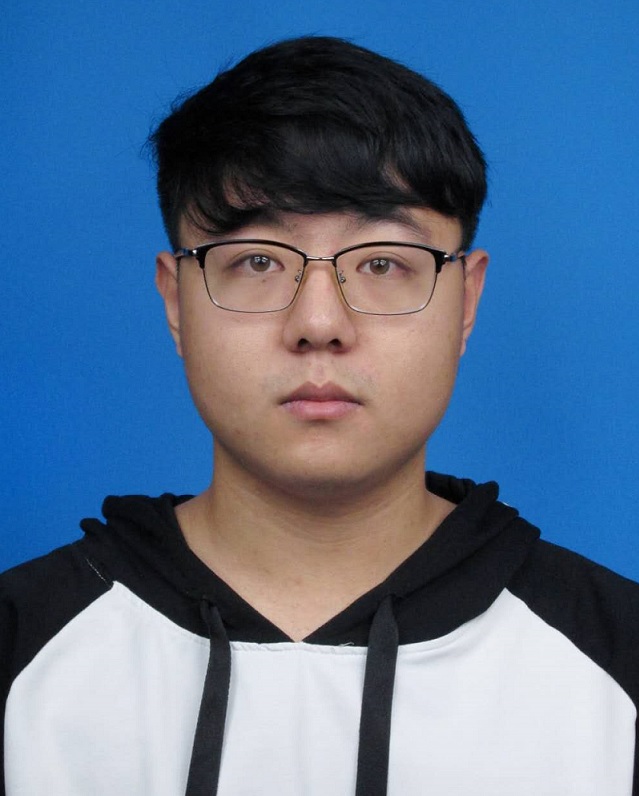}}]{Bochao Liu} is currently an engineer at the Beijing Institute of Astronautical Systems Engineering. He received his Bachelor's degree from the School of Information Science and Engineering, Shandong University, and his Doctoral degree from the School of Cyber Security, University of the Chinese Academy of Sciences. His research interests include differential privacy and information security.
\end{IEEEbiography}

\vspace{-10 mm} 

\begin{IEEEbiography}[{\includegraphics[width=1in,height=1.25in, clip,keepaspectratio]{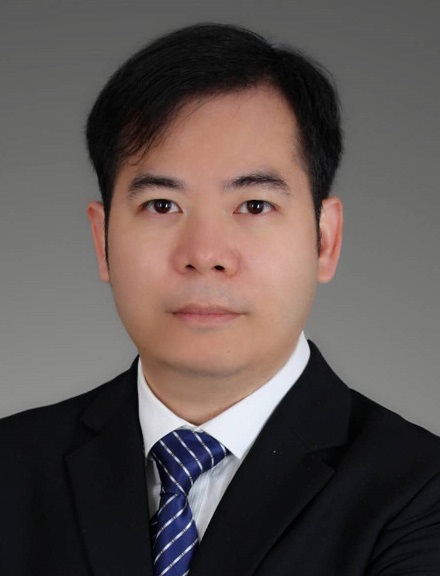}}]{Shiming Ge} (M'13-SM'15) is a professor with the Institute of Information Engineering, Chinese Academy of Sciences. Before that, he was a senior researcher and project manager at Shanda Innovations, a researcher at Samsung Electronics and Nokia Research Center. He received the B.S. and Ph.D. degrees both in Electronic Engineering from the University of Science and Technology of China (USTC). His research mainly focuses on computer vision, data analysis, machine learning and AI safety, especially trustworthy learning solutions towards scalable applications. He has authored and co-authored over 100 research articles including CVPR, NeurIPS, ICML, AAAI, IJCAI, ACM MM, ECCV, TOMM, TNNLS, TCSVT, TMM, TIP, IJCV and TPAMI. He is a senior member of IEEE, CSIG and CCF.
\end{IEEEbiography}

\vspace{-10 mm} 

\begin{IEEEbiography}[{\includegraphics[width=1in,height=1.25in, clip,keepaspectratio]{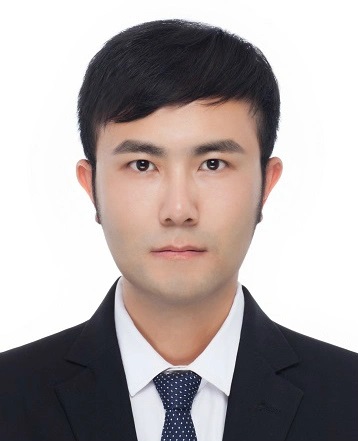}}]{Pengju Wang} is an Assistant Professor with the Institute of Information Engineering, Chinese Academy of Sciences. He received the B.S. degree from the School of Information Science and Engineering at Shandong University, the M.S. degree from the School of Electronic Engineering at Beijing University of Posts and Telecommunications, and the Ph.D. degree from the University of Chinese Academy of Sciences. His research interests include AI safety and federated learning.
\end{IEEEbiography}

\vspace{-10 mm} 

\begin{IEEEbiography}[{\includegraphics[width=1in,height=1.25in, clip,keepaspectratio]{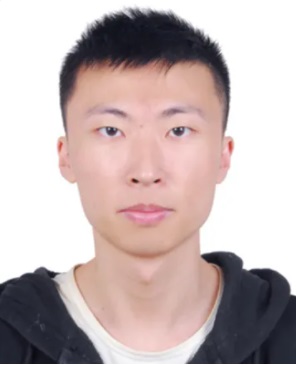}}]
{Shikun Li} is currently a Postdoctoral Research Fellow at the Hong Kong Baptist University. He obtained his Ph.D. degree from the Institute of Information Engineering, Chinese Academy of Sciences in 2024. He has published more than 10 papers at top-tier journals or conferences, including IEEE TPAMI, IEEE TMM, NeurIPS, CVPR, ECCV, AAAI, IJCAI, and ACM MM. His research interests lie in trustworthy machine learning and data-centric AI.
\end{IEEEbiography}

\vspace{-10 mm} 

\begin{IEEEbiography}[{\includegraphics[width=1in,height=1.25in, clip,keepaspectratio]{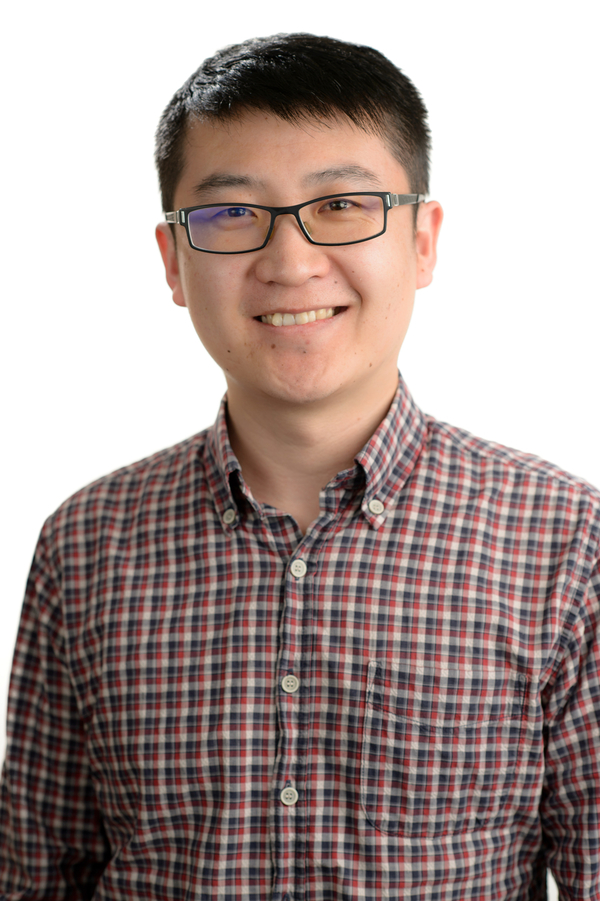}}]{Tongliang Liu} (IEEE Senior Member) is the Director of Sydney AI Centre at the University of Sydney. He is broadly interested in the fields of trustworthy machine learning and its interdisciplinary applications. He has authored and co-authored more than 300 research articles including ICML, NeurIPS, ICLR, CVPR, ICCV, ECCV, AAAI, IJCAI, TPAMI, and JMLR. He is/was a senior meta reviewer for many conferences, such as NeurIPS, ICLR, AAAI, and IJCAI. He is a co-Editor-in-Chief for Neural Networks, an Associate Editor of IEEE TPAMI, IEEE TIP, JAIR, MLJ, TMLR, and ACM Computing Surveys, and is on the Editorial Boards of JMLR. He is a recipient of CORE Award for Outstanding Research Contribution in 2024, the IEEE AI’s 10 to Watch Award in 2022, the Future Fellowship Award from Australian Research Council (ARC) in 2022, the Top-40 Early Achievers by The Australian in 2020, and the Discovery Early Career Researcher Award (DECRA) from ARC in 2018.
\end{IEEEbiography}

\newpage
\onecolumn

\begin{appendices}
\section{Preliminaries}
Here, we overview several fundamental definitions and theorems related to differential privacy. 
\begin{definition}[Differential Privacy (DP)~\cite{dwork2016jpc}]\label{def:dp}
\itshape{A randomized mechanism $\mc{A}$ with domain $\mc{R}$ is $(\varepsilon, \delta)$-DP, if for any subset $\mc{O} \subseteq \mc{R}$ and any adjacent datasets $\mc{D}$ and $\mc{D}'$ :}
\begin{equation}
\begin{aligned}
Pr[\mc{A}(\mc{D}) \in \mc{O}] \leq e^\varepsilon \cdot Pr\left[\mc{A}\left(\mc{D}'\right) \in \mc{O}\right]+\delta,
\end{aligned}
\end{equation}
\end{definition} 
where $\mc{D}$ and $\mc{D}'$ differ from each other with only one example, $\delta \geq 0$ is the failure probability, and $\varepsilon\geq 0$ is the privacy budget. The smaller $\varepsilon$ indicates the better privacy protection. Similarly, label differential privacy (LabelDP)~\cite{ghazi2021nips} is specially defined to protect label privacy. It meets the exact criteria as stated in Def.~\ref{def:dp}, except for the distinction that it applies to any two datasets that differ from each other only in the label of a single example.

\begin{definition}[R$\acute{\textbf{e}}$nyi Differential Privacy (RDP)~\cite{mironov2017renyi}]\label{def:rdp}
\itshape{A randomized mechanism $\mc{A}$ is $(q, \varepsilon)$-RDP with $q > 1$ if for any adjacent datasets $\mc{D}$ and $\mc{D}'$ :}
\begin{equation}
\begin{aligned}
D_{q}&(\mc{A}(\mc{D})||\mc{A}(\mc{D}'))=\frac{1}{q-1}\log \mathbb{E}_{(x\sim\mc{A}(\mc{D}))}\left[\left(\frac{\operatorname{Pr}[\mc{A}(\mc{D})=x]}{\operatorname{Pr}[\mc{A}(\mc{D}')=x]}\right)^{q-1}\right]\le \varepsilon.
\end{aligned}
\end{equation}
\end{definition}
RDP is a more computationally tractable method in calculating privacy, which has a more friendly composition theorem: For a sequence of mechanisms $\{\mc{A}_i\}_{i=1}^k$, where $\mc{A}_i$ is $(q, \varepsilon_i)$-RDP, the composition of them $\mc{A}_1\circ...\mc{A}_i...\circ\mc{A}_k$ is $(q,\sum_i\varepsilon_i)$-RDP. $\mc{A}_i$ represents one query to the teacher in our case. Moreover, the connection between RDP and DP can be described as:
\begin{theorem}[Convert RDP to DP~\cite{mironov2017renyi}]\label{th:rdp-dp}
\itshape{A $(q, \varepsilon)$-RDP mechanism $\mc{A}$ also satisfies $(\varepsilon+\log \frac{q-1}{q}-\frac{\log \delta + \log q}{q-1}, \delta)$-DP.}
\end{theorem}
    
    To provide DP and LabelDP guarantees, we exploit the post-processing~\cite{dwork2014algorithmic} described as follows:
    \begin{theorem}[Post-processing~\cite{dwork2014algorithmic}]\label{th:post-processing}
    \itshape{If mechanism $\mc{A}$ satisfies $(\varepsilon, \delta)$-DP or $(\varepsilon, \delta)$-LabelDP, the composition of a data-independent function $\mc{F}$ with $\mc{A}$ also satisfies $(\varepsilon, \delta)$-DP or $(\varepsilon, \delta)$-LabelDP.}
    \end{theorem}   
DP protects the privacy of both images~(features) and labels, while LableDP protects the privacy of only labels.
To analyze the differential privacy bound for our DPSD, we use Gaussian mechanism~\cite{dwork2014algorithmic,mironov2017renyi} which is achieved by Eq.~\eqref{eq:dp}.
    \begin{theorem}[Gaussian Mechanism]\label{th:gaussian}
    \itshape{Let $f$ be a function with sensitive being $S_f=\max\limits_{\mc{D},\mc{D}^{\prime}}||f(\mc{D})-f(\mc{D}^{\prime})||_2$ over all adjacent datasets $\mc{D}$ and $\mc{D}^{\prime}$. The Gaussian mechanism $\mathcal{A}$ with adding noise to the output of $f$:$\mathcal{A}(x) = f(x) + \mathcal{N}(0, \sigma^2)$
    is $(q, \frac{q S_f^{2}}{2\sigma^2})$-RDP.}
    \end{theorem}
    
\begin{lemma}\label{lemma:sensitive}
    \itshape{For any neighboring gradient vectors $\mathcal{G}, \mathcal{G}^{\prime}$ differing by the gradient vector of one data with length $c$, the $l_2$ sensitivity is $2\beta\sqrt{c}$ after performing normalization with normalization bound $\beta$.}
    \begin{proof}
    \itshape{The $l_2$ sensitivity is the max change in $l_2$ norm caused by the input change. For the vectors after normalization with normalization bound $\beta$, each dimension has a maximum value of $\beta$ and a minimum value of $-\beta$. In the worst case, the difference of one data makes the gradient of all dimensions change from the maximum value $\beta$ to the minimum value $-\beta$, the change in $l_2$ norm equals $\sqrt{(2\beta)^2c}=2\beta\sqrt{c}$.}
    \end{proof}
\end{lemma}

For label-sensitive privacy protection, our approach is a variant of the $\mb{R}(\cdot)$ (Eq.~\eqref{eq:rr}). We first show that $\mb{R}(\cdot)$ is $\varepsilon$-LabelDP.
    \begin{lemma}\label{lemma:rr-data}
    \itshape{For any two labels $y_i$ and $y_j$, $y_i\neq y_j$, $\mb{R}(\cdot)$ satisfies
    \begin{equation}
        \begin{aligned}
            \operatorname{Pr}\left[\mb{R}(y_i; c, \varepsilon)=y\right]\leq e^{\varepsilon} \cdot \operatorname{Pr}\left[\mb{R}(y_j; c, \varepsilon)=y\right],
        \end{aligned}
    \end{equation}
    where $\varepsilon$ and $c$ are same as Eq.~\eqref{eq:rr}.
    }
    \begin{proof}
    \itshape{We know that the maximum of the probability that $\mb{R}(\cdot)$ takes as input $y_i$ and returns $y$ is $e^{\varepsilon}/(e^{\varepsilon} + c - 1)$. The minimum of the probability that $\mb{R}(\cdot)$ takes as input $y_j$ and returns $y$ is $1/(e^{\varepsilon}+c-1)$. Hence, $\operatorname{Pr}\left[\mb{R}(y_i; c, \varepsilon)=y\right]\leq e^{\varepsilon} \cdot \operatorname{Pr}\left[\mb{R}(y_j; c, \varepsilon)=y\right]$.}
    \end{proof}
    \end{lemma}

    \begin{lemma}\label{lemma:rr-dataset}
        \itshape{$\mb{R}(\cdot)$ satisfies $\varepsilon$-LabelDP.}
        \begin{proof}
            For any two label-adjacent datasets $(X,Y)$ and $(X,Y^{\prime})$ that differ only by one label, we simplify them as $Y$ and $Y^{\prime}$ for convenience, since each data is independent of each other, we have
            \begin{equation}\label{eq:indpendent}
                \begin{aligned}
                    \operatorname{Pr}[\mb{R}(Y;m,\varepsilon)\subseteq \mathcal{O}]=\prod_i \operatorname{Pr}[\mb{R}(y_i;m,\varepsilon)=o_i].
                \end{aligned}
            \end{equation}
Assuming $Y=\bar{Y}\bigcup y_i$ and $Y^{\prime}=\bar{Y}\bigcup y_j$, and applying Lemma~\ref{lemma:rr-data} to Eq.~\eqref{eq:indpendent}, we derive
            \begin{equation}
                \begin{aligned}
                    \operatorname{Pr}[\mb{R}(Y;m,\varepsilon)\subseteq \mathcal{O}]
                    &= \operatorname{Pr}[\mb{R}(\bar{Y};m,\varepsilon)\subseteq \mathcal{O}]\cdot \operatorname{Pr}[\mb{R}(y_i;m,\varepsilon)=o]\\
                    &\leq \operatorname{Pr}[\mb{R}(\bar{Y};m,\varepsilon)\subseteq \mathcal{O}]\cdot e^{\varepsilon} \cdot\operatorname{Pr}[\mb{R}(y_j;m,\varepsilon)=o]= e^{\varepsilon}\cdot\operatorname{Pr}[\mb{R}(Y;m,\varepsilon)\subseteq \mathcal{O}].
                \end{aligned}
            \end{equation}
        \end{proof}
    \end{lemma}

\section{Privacy Analysis}
\subsection{The proof of Theorem~\ref{th:dp-theorem}}

\begin{proof}
For each data, the gradient normalization and noise addition implements a Gaussian mechanism which guarantees $(q, \frac{2\beta^2cq}{\sigma^2})$-RDP (Theorem~\ref{th:gaussian} \& Lemma~\ref{lemma:sensitive}). So the data-sensitive privacy protection in our DPSD satisfies $(q, \frac{2\beta^2cbTq}{\sigma^2})$-RDP~(Theorem~\ref{th:post-processing} \& Composition of RDP), which is $(\frac{2\beta^2cbTq}{\sigma^2}+\log \frac{q-1}{q}-\frac{\log \delta + \log q}{q-1},\delta)$-DP~(Theorem~\ref{th:rdp-dp}).
\end{proof}
\vspace{-10pt}
\subsection{Proof of Theorem~\ref{th:theorem-label}}
\begin{proof}
The privacy guarantee of label-sensitive privacy protection in our DPSD follows immediately from that of Lemma~\ref{lemma:rr-dataset} since our choice of k does not depend on the private label or any sensitive information.
\end{proof}

\section{Convergence Analysis}
\subsection{Data-Sensitive Privacy Protection}
Most of our analysis process refers to~\cite{bu2022automatic} and follows the standard assumptions in the SGD literature~\cite{bottou2018optimization}, with an additional assumption on the gradient noise. We assume that {$\mathcal{L}_s$} has a lower bound {$\mathcal{L}_s^{(*)}$ and is $\kappa$-smoothness, described as: $\forall \bl{y},\bl{y}'$}, there is an non-negative constant $\kappa$ such that {$\mathcal{L}_s(\bl{y};\bl{w}_s)-\mathcal{L}_s(\bl{y}';\bl{w}_s) \le \nabla \mathcal{L}_s(\bl{y};\bl{w}_s)^{\top}(\bl{y}-\bl{y}') + \frac{\kappa}{2}||\bl{y}-\bl{y}'||^2$}. The additional assumption is that $({\bl{g}_r-\bl{g}})\sim \mathcal{N}(0,\zeta^2)$, where ${\bl{g}}_r$ is the ideal gradients of {$\mathcal{L}_s$} and ${\bl{g}}$, which we compute, is an unbiased estimate of ${\bl{g}}_r$. For convenience, we set $\mathcal{P}_{\beta,\sigma}({\bl{g}})=\frac{1}{b}\sum\limits^{b}\limits_{i=1}(\frac{\beta\cdot (\mc{M}_k({\bl{g}}))_i}{||\mc{M}_k({\bl{g}})||_2+h} + \mc{N}(0,\sigma^2\beta^2))$. Because {$\mathcal{L}_s$} is $\kappa$-smoothness, we have
    \begin{equation}\label{eq:delta-LT}
    \begin{aligned}
    {\mathcal{L}_s^{(t+1)}-\mathcal{L}_s^{(t)}\leq (\bl{g}_r^{(t)})^{\top}\left(\bl{y}_s^{(t+1)}-\bl{y}_s^{(t)}\right)+\frac{\kappa}{2}||\bl{y}_s^{(t+1)}-\bl{y}_s^{(t)}||^2=-\gamma_s(\bl{g}_r^{(t)})^{\top}}\mathcal{P}_{\beta,\sigma}({\bl{g}})+\frac{\kappa\gamma_s^2}{2}||\mathcal{P}_{\beta,\sigma}({\bl{g}})||^2.
    \end{aligned}
    \end{equation}
    Given $\bl{y}_s^{(t)}$, we can calculate the expectation of {$\mathcal{L}_s^{(t+1)}-\mathcal{L}_s^{(t)}$} as follows
    \begin{equation}\label{eq:Edelta-LT}
    \begin{aligned}
    \mathbb{E}({\mathcal{L}_s^{(t+1)}-\mathcal{L}_s^{(t)}|\bl{y}_s^{(t)})=-\gamma_s(\bl{g}_r^{(t)})^{\top}}\mathbb{E}(\mathcal{P}_{\beta,\sigma}({\bl{g}}))+\frac{\kappa\gamma_s^2}{2}\mathbb{E}(||\mathcal{P}_{\beta,\sigma}({\bl{g}})||^2).
    \end{aligned}
    \end{equation}
    Given the fact that $||\frac{\beta\cdot {\bl{g}}_i}{||{\bl{g}}||_2+e}||^2\leq \beta^2$. We substitute Eq.~\eqref{eq:dp} and combine it with the Cauchy Schwartz inequality to obtain
    \begin{equation}\label{eq:KS}
    \begin{aligned}
    \mathbb{E}(||\mathcal{P}_{\beta,\sigma}({\bl{g}})||^2)\leq 2\beta^2+2\sigma^2\beta^2d,
    \end{aligned}
    \end{equation}
    where $d=||z||^2$, $z\sim\mathcal{N}(0,\text{I}^2)$. So we have
    \begin{equation}\label{eq:Edelta-LT2}
    \begin{aligned}
    \mathbb{E}({\mathcal{L}_s^{(t+1)}-\mathcal{L}_s^{(t)}|\bl{y}_s^{(t)})\leq-\gamma_s \beta(\bl{g}_r^{(t)})^{\top}}\mathbb{E}\left(\frac{{\bl{g}}}{||{\bl{g}}||_2+h}\right)+\kappa\gamma_s^2(\beta^2+\sigma^2\beta^2d).
    \end{aligned}
    \end{equation}
    According to Lemma C.1 in~\cite{bu2022automatic}, we {define $f(c,r;\bl{z})=\frac{(1+rc)}{\sqrt{r^2+2rc+1}+\bl{z}}+\frac{(1-rc)}{\sqrt{r^2-2rc+1}+\bl{z}}$ and} can obtain
    \begin{equation}\label{eq:lemmac1}
    \begin{aligned}
    ({\bl{g}}_r^{(t)})^{\top}\mathbb{E}\left(\frac{{\bl{g}}}{||{\bl{g}}||_2+h}\right)\geq \min _{0<c \leq 1} f\left(c, r ; \frac{h}{\left\|{\bl{g}}_r\right\|}\right) \cdot\left(\left\|{\bl{g}}_r\right\|-\zeta / r\right).
    \end{aligned}
    \end{equation}
    We define $\mathcal{G}(||{\bl{g}}_r||;r;\zeta;h)=\min_{0<c \leq 1} f\left(c, r ; \frac{h}{\left\|{\bl{g}}_r\right\|}\right) \cdot\left(\left\|{\bl{g}}_r\right\|-\zeta / r\right)$. According to the first assumption, it is obtained that
    \begin{equation}\label{eq:l0-l}
    \begin{aligned}
    {\mathcal{L}_s^{(0)}-\mathcal{L}_s^{(*)}}&{\geq\mathcal{L}_s^{(0)}-\mathbb{E}(\mathcal{L}_s)=\sum_t\mathbb{E}(\mathcal{L}_s^{(t)}-\mathcal{L}_s^{(t+1)})}\geq \gamma_s \beta\mathbb{E}\left(\sum_t(\mathcal{G}(||{\bl{g}}_r^{(t)}||))\right)/2-T\kappa \gamma_s^2(\beta^2+\sigma^2\beta^2d).
    \end{aligned}
    \end{equation}
    \begin{equation}\label{eq:E(g)}
    \begin{aligned}
    \mathbb{E}\left(\frac{1}{T}\sum_t\mathcal{G}(||{\bl{g}}_r^{(t)}||)\right)\leq \frac{2(\mathcal{L}_s^{(0)}-\mathcal{L}_s^{(*)})+2T\kappa(\beta^2+\sigma^2\beta^2d)\gamma_s^2}{\beta T\gamma_s}.
    \end{aligned}
    \end{equation}
    Based on the definition of the {top-$k$ selection operator $\mathcal{M}_k$} above, we have
    \begin{equation}
    \begin{aligned}
    \mathcal{G}^{-1}(\bl{z} ; r, \zeta, h)=
    \frac{-\frac{\zeta}{r} h+\left(r^2-1\right) \frac{\zeta}{r} \bl{z}+r h \bl{z}+h \sqrt{\left(\frac{\zeta}{r}\right)^2+2 \zeta \bl{z}+2 h \bl{z}+\bl{z}^2}}{2 h-\left(r^2-1\right) \bl{z}}.
    \end{aligned}
    \end{equation}
    When $r>1$, $\mathcal{G}^{-1}$ doesn't affect the monotonicity of the variable $\bl{z}$. We define $\mathcal{F}(\sqrt{\bl{z}}; r, \zeta, h)=\mathcal{G}^{-1}(\bl{z}; r, \zeta, h)$ and have
    \begin{equation}
    \begin{aligned}
    \min_{0 \leq t \leq T} \mathbb{E}\left(\left\|{\bl{g}}_{r}^t\right\|\right) \leq \mathcal{F}\left( \sqrt{\frac{2\left({\mathcal{L}_s^{(0)}-\mathcal{L}_s^{(*)}}\right)+2T\kappa \gamma_s^2\beta^2(1+\sigma^2d)}{T\gamma_s \beta}} ;\zeta, h\right).
    \end{aligned}
    \end{equation}
    We simply set the learning rate {$\gamma_s \propto 1/{\sqrt{T}}$} and the gradients will gradually tend to 0 as $T$ increases.
\subsection{Label-Sensitive Privacy Protection}
We specifically consider the worst-case scenario wherein the gradient generated by the randomized response mechanism differs completely when the outputs and inputs vary. In this case, applying a randomized response mechanism to the label effectively becomes equivalent to its application on the gradient. Adhering to the same \textbf{five assumptions} in~\cite{bottou2018optimization}, we posit that: (i) $||\nabla \Phi_s(\cdot;\bl{w}_s)-\nabla \Phi_s(\cdot;\bl{w}_s^{\prime})||_2\leq \kappa||\bl{w}_s-\bl{w}_s^{\prime}||_2$; (ii) $\Phi_s(\cdot;\bl{w}_s) \geq \Phi_s(\cdot;\bl{w}_s^{\prime}) + \nabla\Phi_s(\cdot;\bl{w}_s^{\prime})^{{\top}}(\bl{w}_s-\bl{w}_s^{\prime})+\frac{1}{2}c||\bl{w}_s-\bl{w}_s^{\prime}||_2^2$; (iii) $\nabla \Phi_s(\cdot;\bl{w}_s)^{{\top}}\mathbb{E}_{\bl{x}}[{\bl{g}}(\bl{x};\bl{w}_s)]\geq \mu||\nabla\Phi_s(\bl{x};\bl{w}_s)||_2^2$; (iv) $||\mathbb{E}_{\bl{x}}[{\bl{g}}(\bl{x};\bl{w}_s)]||_2\leq \mu_g||\nabla\Phi_s(\bl{x};\bl{w}_s)||_2$; and (v) $\mathbb{V}_{\bl{x}}[{\bl{g}}(\bl{x};\bl{w}_s)]\leq \mu_c + \mu_v||\nabla\Phi_s(\bl{x};\bl{w}_s)||_2^2$, where $\nabla\Phi_s(\bl{x};\bl{w}_s)$ is the true gradient, ${\bl{g}}(\bl{x};\bl{w}_s)$ is the gradient we computed, $\mathbb{E}[\cdot]$ is for mean calculation, $\mathbb{V}[\cdot]$ is for variance calculation and $\kappa,c,\mu,\mu_g,\mu_v,\mu_c$ are non-negative constants. 

    \begin{lemma}\label{lemma:lipschitz}
        \itshape{For any two weights $\bl{w}$ and $\bl{w}^{\prime}$, the difference of the objective function $\Phi_s(\cdot;\bl{w})-\Phi_s(\cdot;\bl{w}^{\prime})$ is limited by the distance between the weights.
        \begin{equation}
            \begin{aligned}
                \Phi_s(\cdot;\bl{w})\leq \Phi_s(\cdot;\bl{w}^{\prime})+\nabla\Phi_s(\cdot;\bl{w}^{\prime})^{{\top}}(\bl{w}-\bl{w}^{\prime})+\frac{1}{2}\kappa ||\bl{w}-\bl{w}^{\prime}||_2^2.
            \end{aligned}
        \end{equation}
        }
        \begin{proof}
            Consider any path $s$ from $\bl{w}^{\prime}$ to $\bl{w}$, according to assumption ({i}), we have
            \begin{equation}
                \begin{aligned}
                    \Phi_s(\cdot;\bl{w}) - \Phi_s(\cdot;\bl{w}^{\prime})&= \int_s\nabla\Phi_s(\cdot;\bl{x})^{{\top}}d\bl{x}=\int_0^1\frac{\partial \Phi_s(\cdot;s(t))}{\partial t}dt=\int_{\bl{w}^{\prime}}^{\bl{w}}\nabla\Phi_s(\cdot;s(t))ds(t)\\
&=\int_{\bl{w}^{\prime}}^{\bl{w}}\nabla\Phi_s(\cdot;\bl{w}^{\prime})ds(t)+\int_{\bl{w}^{\prime}}^{\bl{w}}[\nabla\Phi_s(\cdot;s(t))-\nabla\Phi_s(\cdot;\bl{w}^{\prime})]ds(t)\\
                    &\leq \int_{\bl{w}^{\prime}}^{\bl{w}}\nabla\Phi_s(\cdot;\bl{w}^{\prime})ds(t) + \int_{\bl{w}^{\prime}}^{\bl{w}}\kappa||s(t)-\bl{w}^{\prime}||_2ds(t)=\nabla\Phi_s(\cdot;\bl{w}^{\prime})^{{\top}}(\bl{w}-\bl{w}^{\prime}) +\frac{1}{2}\kappa||\bl{w}-\bl{w}^{\prime}||_2^2.
                \end{aligned}
            \end{equation}
        \end{proof}
    \end{lemma}
    \begin{lemma}\label{lemma:as2}
        \itshape{For any weight $\bl{w}$, the distance between $\Phi_s(\cdot;\bl{w})$ and the minimum value $\Phi_s(\cdot;{\bl{w}^{(*)}})$ is limited by $\nabla \Phi_s(\cdot;\bl{w})$ as:
        \begin{equation}
            \begin{aligned}
                \Phi_s(\cdot;\bl{w})-\Phi_s(\cdot;{\bl{w}^{(*)}})\leq \frac{1}{2c}||\nabla\Phi_s(\cdot;\bl{w})||_2^2.
            \end{aligned}
        \end{equation}
        }
        \begin{proof}
            According to the assumption ({ii}), we can regard the right side of the inequality as a quadratic function on $\bl{w}$. When $\bl{w}=\bl{w}^{\prime}-\frac{1}{c}\nabla\Phi_s(\cdot;\bl{w}^{\prime})$, it takes the minimum value $\Phi_s(\cdot;\bl{w}^{\prime})-\frac{1}{2c}||\nabla\Phi_s(\cdot;\bl{w}^{\prime})||_2^2$. Substituting it into assumption ({ii}) and let $\bl{w}^{\prime}={\bl{w}^{(*)}}$, we can get Lemma~\ref{lemma:as2}.
        \end{proof}
    \end{lemma}
    According to the assumptions before, we {define gradient $\bl{g}^{(t)}=\bl{g}(\bl{x};\bl{w}_s^{(t)})$ and} consider the update at step {$t$} as 
    \begin{equation}
        \begin{aligned}
            {\bl{w}_s^{(t+1)} = \bl{w}_s^{(t)} - \gamma_s\mb{R}[\bl{g}^{(t)}]},
        \end{aligned}
    \end{equation}
where {randomized response mechanism $\mb{R}[\bl{g}^{(t)}]$} will return {$\bl{g}^{(t)}$} with the probability of $e^{\varepsilon}/(e^{\varepsilon}+k-1)$ and return {$-\bl{g}^{(t)}$} with the probability of $1-e^{\varepsilon}/(e^{\varepsilon}+k-1)$. Based on Lemma~\ref{lemma:lipschitz}, we have

    \begin{equation}
        \begin{aligned}
            \Phi_s(\cdot;{\bl{w}_s^{(t+1)}})\leq \Phi_s(\cdot;{\bl{w}_s^{(t)}})-&\gamma_s\nabla\Phi_s(\cdot;{\bl{w}_s^{(t)}})^{{\top}}\mb{R}[{\bl{g}^{(t)}}]
            +\frac{1}{2}\kappa\gamma_s^2\underbrace{||\mb{R}[{\bl{g}^{(t)}}]||_2^2}_{||{\bl{g}^{(t)}}||_2^2}.
        \end{aligned}
    \end{equation}
    Taking the expectations on both sides gives
    \begin{equation}
        \begin{aligned}
            \mathbb{E}[\Phi_s(\cdot;{\bl{w}_s^{(t+1)}})-\Phi_s(\cdot;{\bl{w}_s^{(t)}})]&\leq-\gamma_s\nabla\Phi_s(\cdot;{\bl{w}_s^{(t)}})^{{\top}}\mathbb{E}[\mb{R}[{\bl{g}^{(t)}}]]+\frac{1}{2}\gamma_s^2\kappa\underbrace{\mathbb{E}[||{\bl{g}^{(t)}}||^2_2]}_{||\mathbb{E}[{\bl{g}^{(t)}}]||_2^2+\mathbb{V}[{\bl{g}^{(t)}}]}.
        \end{aligned}
    \end{equation}
     According to our pre-assumed scenario, 
     \begin{equation}
         \begin{aligned}
             \mathbb{E}[\mb{R}[{\bl{g}^{(t)}}]]&=\frac{e^{\varepsilon}}{e^{\varepsilon}+k-1}{\bl{g}^{(t)}}-(1-\frac{e^{\varepsilon}}{e^{\varepsilon}+k-1}){\bl{g}^{(t)}}=\underbrace{(\frac{2e^{\varepsilon}}{e^{\varepsilon}+k-1}-1)}_{\varsigma}{\bl{g}^{(t)}}.
         \end{aligned}
     \end{equation}
By combining with the assumptions ({iii}), ({iv}) and ({v}), we can get
     \begin{equation}\label{eq:neq}
        \begin{aligned}
            \mathbb{E}[\Phi_s(\cdot;{\bl{w}_s^{(t+1)}})-\Phi_s(\cdot;{\bl{w}_s^{(t)}})] &\leq -\gamma_s\varsigma\nabla\Phi_s(\cdot;{\bl{w}_s^{(t)}})^{{\top}}\mathbb{E}[{\bl{g}^{(t)}}] +\frac{1}{2}\gamma_s^2\kappa(||\mathbb{E}[{\bl{g}^{(t)}}]||_2^2+\mathbb{V}[{\bl{g}^{(t)}}])\\
            &\leq -\gamma_s\varsigma\mu||\nabla\Phi_s(\cdot;{\bl{w}_s^{(t)}})||_2^2+\frac{1}{2}\gamma_s^2\kappa(\mu_c+(\mu_g^2+\mu_v)||\Phi_s(\cdot;{\bl{w}_s^{(t)}})||_2^2)\\
            &=\underbrace{(-\gamma_s\varsigma\mu+\frac{1}{2}\gamma_s^2\kappa(\mu_g^2+\mu_v))}_{\tau}||\Phi_s(\cdot;{\bl{w}_s^{(t)}})||_2^2+\frac{1}{2}\gamma_s^2\kappa \mu_c.
        \end{aligned}
     \end{equation}
     If the algorithm converges, it takes $-\gamma_s\mu+\frac{1}{2}\gamma_s^2\kappa(\mu_g^2+\mu_v) < 0$. According to Lemma~\ref{lemma:as2}, we can further get
     \begin{equation}\label{eq:db}
         \begin{aligned}
             {\mathbb{E}[\Phi_s(\cdot;{\bl{w}_s^{(t+1)}})-\Phi_s(\cdot;{\bl{w}_s^{(t)}})]} & {=
             \mathbb{E}[\Phi_s(\cdot;{\bl{w}_s^{(t+1)}})-\Phi_s(\cdot;\bl{w}_s^{(*)})]-\mathbb{E}[\Phi_s(\cdot;\bl{w}_s^{(t)})-\Phi_s(\cdot;\bl{w}_s^{(*)})]}\\
             &{\leq \tau||\Phi_s(\cdot;{\bl{w}_s^{(t)}})||_2^2+\frac{1}{2}\gamma_s^2\kappa \mu_c \leq 2\tau c\mathbb{E}[\Phi_s(\cdot;{\bl{w}_s^{(t)}})-\Phi_s(\cdot;{\bl{w}_s^{(*)}})]+\frac{1}{2}\gamma_s^2\kappa \mu_c,}
         \end{aligned}
     \end{equation}
{where the first inequation uses Eq.~\eqref{eq:neq} and the second inequation is based on Lemma~\ref{lemma:as2}. Then} Eq.~\eqref{eq:db} is transformed to
    \begin{equation}
        \begin{aligned}
            \mathbb{E}[\Phi_s(\cdot;{\bl{w}_s^{(t+1)}})-\Phi_s(\cdot;{\bl{w}_s^{(*)}})]+\frac{\gamma_s^2\kappa \mu_c}{4\tau c} \leq (2\tau c+1)(\mathbb{E}[\Phi_s(\cdot;{\bl{w}_s^{(t)}})-\Phi_s(\cdot;{\bl{w}_s^{(*)}})]+\frac{\gamma_s^2\kappa \mu_c}{4\tau c}).
        \end{aligned}
    \end{equation}
When guaranteeing $\tau<0$, then $2\tau c + 1<1$ and the algorithm converges. The error from the minimum $\Phi_s(\cdot;{\bl{w}_s^{(*)}})$ is $-\frac{\gamma_s^2\kappa \mu_c}{4\tau c}$.

\section{FedDPSD: DPSD Variant under Federated Learning Setup}\label{appendix:fed}
We present the algorithm details of our DPSD approach under federated learning setup in Alg.~\ref{alg:FedDPSD}. For experiments setup, we adopt the hyperparameters setting in~\cite{fang2022up} for the GAN training and hyperparameters setting in~\cite{zhang2022dense} for the federated setting. 
\begin{algorithm}[h]
    \caption{FedDPSD}
    \label{alg:FedDPSD}
    \textbf{Input}: Client number $m$, local teacher models $\{\Phi^j_t(\bl{w}^j_t;\cdot)\}_{j=1}^m$, training iterations $T$, loss function $\mathcal{L}_s, \mathcal{L}_S, \mathcal{L}_G$, noise scale $\sigma$, sample size $b$, learning rate $\gamma, \gamma_s, \gamma_g$, gradient normalization bound $\beta$, positive stability constant $e$, selection number $k$, randomized response algorithm $\mb{R}_\varepsilon(\cdot)$, uniform sampling algorithm $\mathbb{U}(\cdot)$, binary switch sign $s \in \{0, 1\}$\\
    \begin{algorithmic}[1] 
    \FOR {$t \in [T]$}
    \STATE /* Server */
    \STATE Sample $b$ noise samples $\textbf{z}=\{z_i\}_{i=1}^{b}$
    \STATE Generate $b$ synthetic samples $\mc{D}=\{\bl{x}_i=\Phi_g(\bl{w}_g;z_i)\}_{i=1}^{b}$
    \STATE Compute the predictions $\Phi_s(\mc{D};\bl{w}_s)$
    \STATE Send the data $\mc{D}$ and predictions $\Phi_s(\mc{D};\bl{w}_s)$ to each client
    \STATE /* Clients */
    \FOR{each client $j\in [m]$ in parallel do}
    \STATE Compute the predictions $\Phi^j_t(\mc{D};\bl{w}^j_s)$
    \STATE $\mathcal{O}_j$ = Fed-FlexPer($\Phi^j_t(\mc{D};\bl{w}^j_t)$,$\Phi_s(\mc{D};\bl{w}_s)$)
    \STATE Send $\mathcal{O}_j$ to the server.
    \ENDFOR
    \STATE /* Server */
    \STATE Compute the differentially private labels $\Phi^{\prime}_s(\mc{D}) = s\cdot\frac{1}{m}(\sum\limits^{m}\limits_{i=1}\mathcal{O}_i + \mathcal{N}(0,\sigma^2\beta^2)) + (1-s)\cdot\frac{1}{m}\sum\limits^{m}\limits_{i=1}\mathcal{O}_i$
    \STATE Compute loss $\mathcal{L}_S(\Phi_s(\mc{D};\bl{w}_s), \Phi_s^{\prime}(\mc{D}))$
    \STATE Update student $\bl{w}_s^{(t+1)}=\bl{w}_s^{t}-\gamma_s\cdot \frac{\partial \mathcal{L}_S}{\partial \bl{w}_s^{t}}$
    \STATE Compute loss $\mathcal{L}_G(\Phi_s(\mc{D};\bl{w}_s))$
    \STATE Update generator $\bl{w}_g^{t+1}=\bl{w}_g^{t}-\gamma_g \cdot\frac{\partial (\mathcal{L}_S+\mathcal{L}_G)}{\partial \bl{w}_g^{t}}$
    \ENDFOR
    \STATE \textbf{return} $\bl{w}_s$ and $\bl{w}_g$\\
    
    \STATE\textbf{Function} Fed-FlexPer($\Phi_t(\mc{D};\bl{w}_t),\Phi_s(\mc{D};\bl{w}_s)$)\\
    \STATE /* Compute the data-protected gradients $\bar{g}$ */
    \FOR {($\Phi_t(\bl{x}_i;\bl{w}_t),\Phi_s(\bl{x}_i;\bl{w}_s)$) in ($\Phi_t(\mc{D};\bl{w}_t),\Phi_s(\mc{D};\bl{w}_s)$)}
    \STATE Compute loss $\mathcal{L}_s(\Phi_t(\bl{x}_i;\bl{w}_t),\Phi_s(\bl{x}_i;\bl{w}_s))$
    \STATE Compute the gradient $g_i=\frac{\partial \mathcal{L}_s}{\partial \Phi_s(\bl{x}_i\bl{w}_s)}$
    \STATE Normalize the gradient $\bar{g}_i=\frac{\beta\cdot g_i}{||g||_2+e}$
    \ENDFOR
    \STATE /* Compute the label-protected labels $\bl{y}_l$ */
    \FOR {each synthetic data $\bl{x}_i$ in $\mc{D}$}
    \STATE Select the set of indexes of top-$k$ values $\mc{I}_i$ according $\Phi_s(\bl{x}_i;\bl{w}_s)$
    \STATE Determine whether or not $\arg\max(\Phi_t(\bl{x}_i;\bl{w}_t))$ is in $\mc{I}_i$
    \STATE $\bl{y}_l[i]=\mb{R}_\varepsilon(\mc{I}_i;\Phi_t(\bl{x}_i,\bl{w}_t))$ if so and $\bl{y}_l[i]=\mb{U}(\mc{I}_i)$ if not
    \ENDFOR
    \STATE /* Ensemble outputs with switch $s$ */
    \STATE \textbf{return} $s\cdot\bar{g} + (1-s)\cdot \bl{y}_l$
\end{algorithmic}
\end{algorithm}




\end{appendices}

\end{document}